\documentclass[preprint,12pt,authoryear]{elsarticle}

\usepackage{amssymb}

\usepackage{lineno}

\usepackage{hyperref}       
\usepackage{url}            
\usepackage{booktabs}       
\usepackage{amsfonts}       
\usepackage{nicefrac}       
\usepackage{microtype}      
\usepackage{fancyhdr}       
\usepackage{graphicx}       
\usepackage{xcolor}
\usepackage{amsmath}
\usepackage{amsfonts}
\usepackage{caption}
\usepackage{multirow}
\usepackage{subcaption}
\usepackage{bbm}
\usepackage{float}
\usepackage{pifont}
\newcommand{\cmark}{\ding{51}}%
\newcommand{\xmark}{\ding{55}}%

\journal{Information Fusion}

\begin{document}

\begin{frontmatter}



\title{Interpretable Prognostics with Concept Bottleneck Models}
\author{Florent Forest}
\author{Katharina Rombach}
\author{Olga Fink}

\affiliation{organization={Intelligent Maintenance and Operations Systems, EPFL},
            city={Lausanne},
            country={Switzerland}}

\begin{abstract}
Deep learning approaches have recently been extensively explored for the prognostics of industrial assets. However, they still suffer from a lack of interpretability, which hinders their adoption in safety-critical applications. To improve their trustworthiness, explainable AI (XAI) techniques have been applied in prognostics, primarily  to quantify the importance of input variables for predicting the remaining useful life (RUL) using post-hoc attribution methods. In this work, we propose the application of Concept Bottleneck Models (CBMs), a family of inherently interpretable neural network architectures based on concept explanations, to the task of RUL prediction. Unlike attribution methods, which explain decisions in terms of low-level input features, concepts represent high-level information that is easily understandable by users. Moreover, once verified in actual applications, CBMs enable  domain experts to intervene on the concept activations at test-time. We propose using the different degradation modes of an asset as intermediate concepts. Our case studies on the New Commercial Modular AeroPropulsion System Simulation (N-CMAPSS) aircraft engine dataset for RUL prediction demonstrate that the performance of CBMs can be on par or superior to black-box models, while being more interpretable, even when the available labeled concepts are limited. Code available at \href{https://github.com/EPFL-IMOS/concept-prognostics/}{\url{github.com/EPFL-IMOS/concept-prognostics/}}.

\end{abstract}


\end{frontmatter}


\section{Introduction}

Prognostics is the task of forecasting how long an industrial asset will be able to operate and fulfill its function after the onset of a fault \citep{biggio_prognostics_2020}. As an integral part of the Prognostics and Health Management (PHM) engineering field, it is a crucial enabler of predictive maintenance strategies and improves the availability, safety and cost-efficiency of industrial assets. Based on condition monitoring data, prognostics aims to predict the remaining useful life (RUL) of an asset \citep{lei_machinery_2018}. Methods to estimate the RUL can be broadly categorized into physics-based, statistical model-based approaches, and data-driven approaches \citep{dersin_modeling_2023}. In recent years, data-driven approaches using  deep learning, i.e., employing deep neural networks to predict the RUL from raw condition monitoring signals, have been explored in the literature, achieving promising performance \citep{fink_potential_2020}. Many works have used recurrent neural networks (RNN) for RUL prediction \citep{heimes_recurrent_2008}, particularly the long short-term memory cell (LSTM) \citep{zheng_long_2017,huang_bidirectional_2019,shi_dual-lstm_2021} and the gated recurrent unit (GRU) \citep{chen_gated_2019,baptista_using_2020}. Alternatively, convolutional neural networks (CNN) \citep{li_remaining_2018,zhu_estimation_2019,arias_chao_fusing_2022} process the condition monitoring signals in a sliding window manner. Recently, Transformer models based on the attention mechanism have also been applied for this task \citep{zhang_dual-aspect_2022,biggio_ageing-aware_2023,xiang_bayesian_2024}.

While deep learning has the advantage of automatically learning features from large amounts of high-dimensional condition monitoring signals and achieving high predictive performance, these approaches also face challenges. 
In particular, the lack of interpretability of the predictions from black-box models hinders their adoption in safety-critical contexts, as domain experts need to understand the predictions of their models before taking concrete maintenance or operational actions. Note that the terms \emph{interpretable} and \emph{explainable} will be used interchangeably in this work \citep{lipton_mythos_2017}. 
In this regard, the field of eXplainable Artificial Intelligence (XAI) aims to make AI systems more interpretable and trustworthy while maintaining their performance \citep{arrieta_explainable_2019}. Its application in the context of PHM, predictive maintenance and digital twins has received growing interest over the recent years \citep{nor_overview_2021,vollert_interpretable_2021,cummins_explainable_2024}. In particular, the necessity to interpret prognostics models has been highlighted as a priority in recent research studies \citep{serradilla_interpreting_2020,baptista_relation_2022,solis-martin_soundness_2023}. Generally, techniques in XAI can be categorized into two main types of approaches. The first type is \emph{post-hoc} interpretability, where the decisions of an already trained black-box model are explained \emph{a posteriori}. The second category involves interpretability that is directly built into the model, making it interpretable \emph{by-design}. This approach is also  referred to as \emph{intrinsic} interpretability.

Due to their flexibility and ease-of-use, post-hoc methods have already been widely applied to prognostics tasks in the literature. Among these, feature attribution methods are indirect ways of explaining a model by calculating an importance score for each variable or feature handled by the model (such as pixels in an image, time points in a signal, or frequencies in a spectrum) to predict the target outcome. 
For instance, the importance of each input variable in predicting the RUL can be quantified using SHAP values \citep{hong_remaining_2020,khan_explainable_2022,youness_explainable_2023}, gradient-weighted class attribution maps (Grad-CAM) \citep{solis-martin_soundness_2023} or LIME \citep{baptista_using_2020,protopapadakis_explainable_2022,khan_explainable_2022}. However, the authors observed limited fidelity of LIME's local linear explanations. While the aforementioned techniques provide explanations in a local neighborhood around the input, coordinate plots of sets of local linear models can be used to assess feature importance on a global scale \citep{kobayashi_explainable_2024}. Moreover, these techniques may disagree on the importance of features, requiring additional evaluations to select the most trustworthy explanation \citep{kundu_explainable_2023}. 


While attribution explanations are relatively easy to interpret in vision tasks, where they highlight specific regions in the input images, they are more difficult to interpret in complex signals such as multivariate time series from sensors. As seen in the various  studies, attributions can be noisy and scattered, making the  explanations provided not easily understandable, even for domain experts \citep{decker_does_2023}. Moreover, these explanations are based on low-level features that do not correspond to human-understandable concepts. More generally, post-hoc explanations have significant  limitations, as they do not elucidate the model's decision-making process \citep{rudin_stop_2019}, are generally limited to a local neighborhood around the input to be explained, and are not always able to detect reliance on spurious correlations in the data \citep{adebayo_post_2022}. The reliability of post-hoc methods has been questioned in the literature, as they can be sensitive to small changes in the input \citep{kindermans_reliability_2017,ghorbani_interpretation_2018}. 

For all these reasons, exploring strategies for intrinsic interpretability and alternative types of explanations is highly valuable. 
In PHM applications, such intrinsic interpretability can be integrated in several ways. When the degradation mechanisms are known, model-based approaches based on stochastic processes \citep{li_stochastic_2000}, physics \citep{oppenheimer_physically_2002} or expert rules \citep{chen_wind_2013} can be used, which are inherently interpretable. Symbolic regression approaches based on evolutionary computing can also yield transparent RUL regression models \citep{ding_dynamic_2021}. 
However, when the decision is made to apply data-driven methods such as machine learning, knowledge on the degradation processes is typically not available and prevents the use of knowledge-based approaches.
A straightforward way to obtain an interpretable machine learning model is to choose a simple hypothesis class, such as linear models or small decision trees that are inherently directly interpretable. However, this choice involves a trade-off with predictive performance, which may not always be acceptable \citep{veiber_challenges_2020}. A different strategy is to leverage physics knowledge in combination with observed data, i.e., physics-informed machine learning \citep{karniadakis_physics-informed_2021}. Physics-informed neural networks incorporate physical equations in their loss function or embed physical quantities as intermediate or output neurons, making them inherently more interpretable to domain experts. Recently, the application of physics-informed machine learning to PHM \citep{deng_physics-informed_2023} and specifically 
 to prognostics \citep{li_review_2024,bajarunas_generic_2023} has been explored. However, this topic is beyond the scope of this paper. Alternatively, interpretability can be increased by modifying the model architecture or decomposing a complex model in a modular way. For instance, \cite{kraus_forecasting_2019} propose a structured-effect neural network, where the model is decomposed into three components: a distribution-based lifetime component, a linear component and a RNN to model the remaining variability. Interpretability is then increased by the transparent nature of the first two components, as well as uncertainty estimation. A different direction has been taken in \cite{lee_explainable_2020}, where the model's interpretability is increased by filtering and reducing the dimension of latent representations in an autoencoder, allowing for direct feature visualization.

To enable an inherent interpretability of deep learning-based prognostics models, this paper focuses on a family of interpretable network architectures based on \emph{concept explanations}, i.e., explaining a prediction in terms of high-level, human-understandable characteristics called \emph{concepts}. For example, the classification of an image of an apple may be explained by concepts such as shape ("round") and color ("red") \citep{zarlenga_concept_2022}. More specifically, we study neural network models using a \emph{concept bottleneck}, which integrate user-defined concepts as an intermediate layer in their architecture. 
Intermediate concepts, also referred to as \emph{attributes} or \emph{semantic bottlenecks} depending on the context, have been widely applied  across  various fields, particularly  in  computer vision. These concepts serve to improve the model's interpretability \citep{bucher_semantic_2018,losch_interpretability_2019,marcos_semantically_2019} and generalization capabilities \citep{kumar_attribute_2009,lampert_learning_2009,li_deep_2018}.  
Conversely, post-hoc concept analysis methods, such as Testing with Concept Activation Vectors (TCAV) \citep{kim_interpretability_2018}, explain a network in a post-hoc manner by quantifying the sensitivity of the model to a set of user-defined concepts. However, since the model is not specifically trained to learn those concepts, it may fail  to effectively capture them. Furthermore, these methods also do not permit test-time intervention on concepts which can be particularly relevant for practical applications \citep{koh_concept_2020}. 
Concept bottleneck models (CBMs) \citep{koh_concept_2020} first predict a set of concepts in their intermediate bottleneck, and then use those predicted concepts to predict the target. To be trained, each input must be labeled with a set of associated concepts, in addition to the target label for the task at hand. A central hypothesis in this architecture is that the concepts must be sufficient to accurately predict the target. However, since the concepts are user-defined, require labeling, and must be sufficiently high-level  to be human-understandable, their quantity  and expressiveness tend to be limited. This issue  of \emph{incomplete concepts} leads to a trade-off between performance and interpretability, typically resulting in lower performance compared to traditional, non-interpretable deep learning models \citep{yeh_completeness-aware_2022}. Several solutions have been proposed to overcome this limitation, among them Concept Embedding Models (CEM), which enrich the representations of concepts compared with previous CBMs \citep{zarlenga_concept_2022}. Despite their potential, these approaches have not yet been applied to PHM applications.

We propose to apply concept bottleneck models to the task of predicting remaining useful life, considering the different degradation modes of an asset as our concepts. We posit that the RUL can be predicted  based on the current degradation state of each of its components. Our proposed methodology is illustrated through a case study on turbofan aircraft engines. This work introduces concept-based approaches to prognostics for the first time. In addition, at the more general methodological level, to the best of our knowledge, CBMs have not yet been applied to regression tasks, with previous works focusing exclusively on classification. The continuous nature of regression outputs creates an additional challenge, as concepts are defined as discrete variables, but their representations must enable the prediction of a continuous target. A second important challenge of developing CBM approaches for PHM applications is that in real-world industrial settings, concept labels are likely to be scarce and thus incomplete. Nevertheless, our research will demonstrate  that this approach can still achieve robust performance under these constraints, thanks to rich concept representations. We will demonstrate that the performance of these models can match or even surpass that of their non-interpretable counterparts.

One of the notable additional advantages of CBMs is that they allow for human interventions at inference time, called \textit{test-time interventions}, where a human operator can confirm the correctly predicted concepts and correct the mispredicted concepts, possibly changing the CBM's final output \citep{koh_concept_2020}. 
In the task of RUL prediction based on component degradation concepts, we propose to use test-time interventions to improve the predicted RUL value by correcting the predicted degradation state of a given component to its true state at a given time point. 
Performing test-time interventions in prognostics models presents important difficulties compared with standard image classification problems, because observations are not statistically independent, and determining the true degradation state of an industrial asset requires a costly inspection or diagnostic operation. 
Therefore, realistic test-time interventions strategies need to be devised to determine when and where to intervene, which constitutes our second main contribution.

Our contributions can be summarized as follows:
\begin{itemize}
    \item We propose concept bottleneck models for more interpretable prognostics, where degradation modes of an asset are used as intermediate concepts. This approach predicts the remaining useful life in a more human-understandable way compared to standard deep learning approaches.
    \item We conduct a comprehensive evaluation and comparison of several concept-based models through an in-depth case study on turbofan aircraft engine prognostics. We assess these methods based on RUL prediction performance, concept quality, as well as fault detection. In addition, ablation studies are conducted to evaluate the influence of the number of concepts included for training.
    \item We demonstrate that the additional interpretability of concept-based models does not imply a significant loss in performance compared with traditional non-interpretable models.
    \item We propose a test-time concept intervention strategy adapted to prognostics applications.
\end{itemize}

The remainder of the paper is structured as follows. Section~\ref{sec:background} introduces the necessary background on the methods. Section~\ref{sec:method} presents our proposed concept-based architectures for RUL prediction, as well as a test-time intervention strategy for prognostics. Section~\ref{sec:experiments} details experimental settings such as data, evaluation metrics, and methods compared. Section~\ref{sec:results} reports the results of our experiments and studies. Further discussions are elaborated in Section~\ref{sec:discussion}. Finally, Section~\ref{sec:conclusion} concludes this work.

\section{Background}\label{sec:background}





Concept Bottleneck Models (CBMs) \citep{koh_concept_2020} are a family of neural networks that enforce the prediction of human-understandable concepts in an intermediate layer, called the \emph{bottleneck}. In order to be trained, CBMs require a dataset of samples ($\mathbf{x}$, $y$, $\mathbf{c}$), where each input $\mathbf{x}$ is labeled with ground-truth concepts $\mathbf{c}$, in addition to the target label $y$. Given an input $\mathbf{x}$, they first predict concepts $\hat{\mathbf{c}}$, and then use those predicted concepts to predict the target $\hat{y}$. The concepts are generally multi-label, meaning that several concepts can be true for a single sample. Hence, $\mathbf{c}$ is a multi-label binary vector.


The first version of CBMs predicts scalar concept representations in the intermediate bottleneck, i.e., $\mathbf{\hat{c}} = \phi(\mathbf{x}) \in \mathbb{R}^k$, with each dimension of the concept vector aligned with a single ground-truth concept. The concept activations $\hat{c}_1, \cdots, \hat{c}_k$ are obtained by applying an element-wise activation function $s: \mathbb{R} \rightarrow [0, 1]$. Boolean CBMs use a hard threshold activation $s(z) = \mathbbm{1}_{z>0.5}$, while Fuzzy CBMs use a sigmoidal activation $s(z) = 1/(1+e^{-z})$. Then, the output is computed from the predicted concept activations: $\hat{y} = f(s(\hat{\mathbf{c}}))$. Following \cite{koh_concept_2020}, we use joint bottleneck training with a weighting coefficient $\lambda$, resulting in the following loss function for a regression task:

\begin{equation}
    \mathcal{L}_{\text{CBM}} = \text{MSE}(y, \hat{y}) + \lambda \cdot \text{BCE}(\mathbf{c}, s(\mathbf{\hat{c}}))
    \label{eq:loss}
\end{equation}

where the task loss is a mean squared error (MSE) between the predicted and the true target value, and the concept loss is a binary cross-entropy (BCE) between the predicted concept activations and the true concept labels.

In Hybrid CBMs, the scalar concept representations include  $m$ additional unsupervised dimensions, resulting in concept vectors $\hat{\mathbf{c}} \in \mathbb{R}^{k+m}$. This extra capacity improves the task performance by alleviating the issue of concept incompleteness \citep{mahinpei_promises_2021}, but hinders the interpretability of the bottleneck. The activation function is  applied only to the first $k$ dimensions. The output is then computed as $\hat{y} = f([s(\mathbf{\hat{c}}_{1:k}), \hat{\mathbf{c}}_{k+1:k+m}])$, where the brackets represent the concatenation operation. The resulting loss function is expressed as follows:

\begin{equation}
    \mathcal{L}_{\text{HybridCBM}} = \text{MSE}(y, \hat{y}) + \lambda \cdot \text{BCE}(\mathbf{c}, s(\mathbf{\hat{c}}_{1:k})).
\end{equation}

Concept Embedding Models (CEMs) \citep{zarlenga_concept_2022} are an extension of CBMs where concepts are represented by vectors, called \emph{embeddings}, instead of scalar representations. Initially, inputs are passed through a feature extractor $\phi(\cdot)$, resulting in a latent code $\mathbf{z}$. This vector is then processed by the set of functions $(\phi_1^+, \phi_1^-), \cdots, (\phi_k^+, \phi_k^-)$, outputting a positive ($\mathbf{c}^+_i \in \mathbb{R}^m$) and negative ($\hat{\mathbf{c}}^-_i \in \mathbb{R}^m$) embedding corresponding to each concept. This design allows encoding of information that is both positively and negatively related to the activation of each concept. Subsequently, a probability $\hat{p}_i$ for each concept is computed using a linear scoring function that is shared across all concepts, enabling  the computation of a mixture of the positive and negative embeddings, i.e., $\mathbf{\hat{c}}_i = \hat{p}_i \hat{\mathbf{c}}^+_i + (1-\hat{p}_i) \hat{\mathbf{c}}^-_i) \in \mathbb{R}^m$. Otherwise, the loss function is the same as that of other CBMs.

An additional modification compared to CBMs is the introduction of random interventions during training. Instead of  using only the predicted concepts $\mathbf{\hat{c}}$ in  joint bottleneck training, the ground-truth concepts $\mathbf{c}$ are randomly used with some probability. \cite{zarlenga_concept_2022} observed  a performance improvement for test-time interventions when using this technique.

\section{Method}\label{sec:method}

%
\subsection{Component degradation modes as understandable concepts}

\begin{figure}
    \centering
    \begin{subfigure}[b]{\textwidth}
         \centering
         \includegraphics[height=13em]{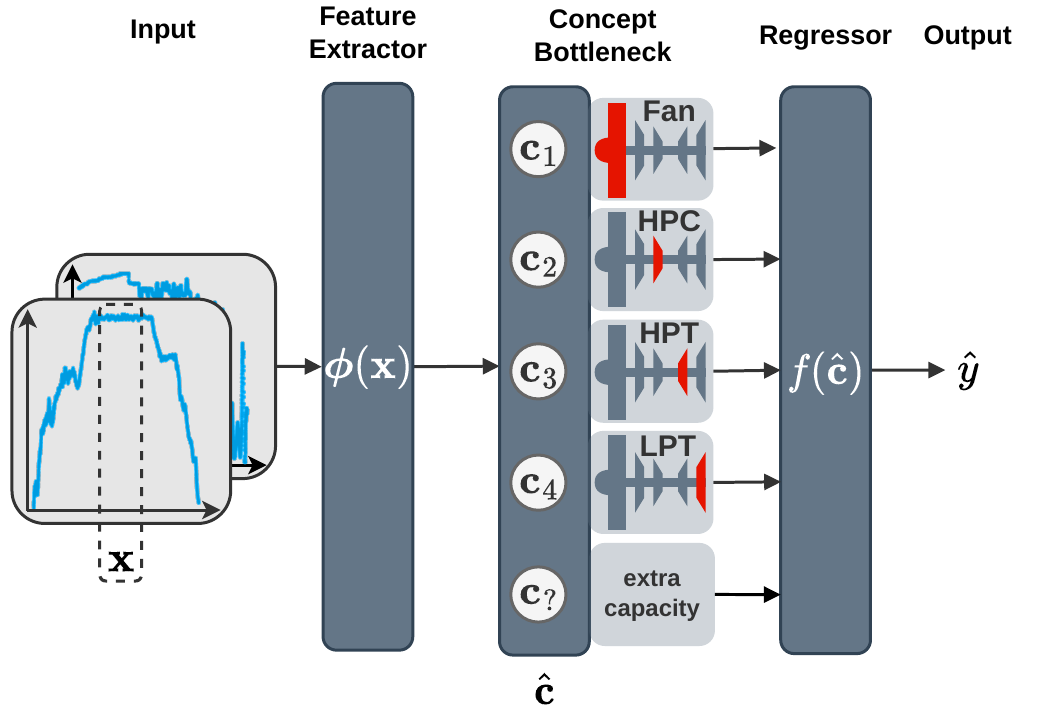}
         \caption{Concept Bottleneck Model (CBM), where $\mathbf{c}_{\text{?}}$ denotes extra representational capacity in case of a hybrid CBM.}
         \label{fig:architecture-cbm}
     \end{subfigure}
    \begin{subfigure}[b]{\textwidth}
         \centering
         \includegraphics[height=13em]{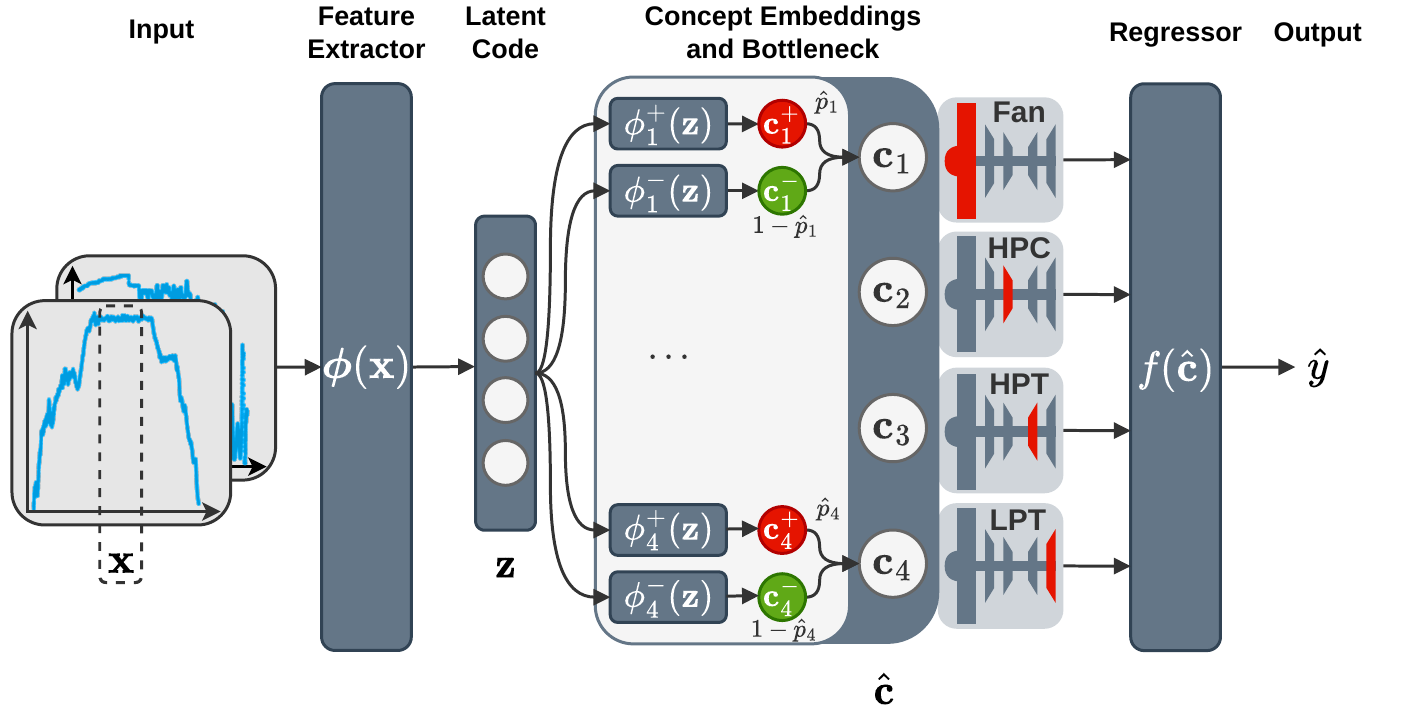}
         \caption{ Concept Embedding Model (CEM), where positive ($\mathbf{c}^+_i$) and negative ($\mathbf{c}^-_i$) embeddings are learned for each concept.}
         \label{fig:architecture-cem}
     \end{subfigure}
    \caption{Our proposed concept-based architectures for interpretable prognostics. Intermediate concepts $\mathbf{c}_i$ corresponding to component degradation modes are used to predict the remaining useful life (in this example, four concepts correspond to the Fan, HPC, HPT and LPT components of a turbofan engine).}
    \label{fig:architectures-concept}
\end{figure}

We propose considering component degradation modes as our intermediate concepts to predict the RUL, defining a distinct concept for each degrading component. An illustration of our methodology is represented in Fig.~\ref{fig:architectures-concept}. The underlying assumption is that the RUL of an asset is determined by the states of degradation of each of its components. For instance, if the asset considered is a turbofan aircraft engine, the components can be defined as the different stages of compressors and turbines. The degradation state of each of the components are then considered as one concept. In addition, if a component is subject to several failure modes, it can be further sub-divided into several sub-concepts. As concepts are defined as binary attributes, we define one binary concept for each degradation mode, indicating whether a given degradation mode is active (1, e.g., the corresponding component is faulty) or inactive (0, e.g., the corresponding component is healthy). Unlike post-hoc attribution methods, whose explanations might not be understandable to humans, these concepts are user-defined and represent high-level pieces of information that are known \textit{a priori} and are human-understandable. In addition, knowing such concepts is particularly beneficial for decision-makers, as it enables them to take more targeted maintenance actions.

Let us consider a dataset ($\mathbf{x}$, $y$, $\mathbf{c}$), where $\mathbf{x}$ denotes the input condition monitoring signals, $y$ the associated RUL value, and $\mathbf{c}$ the degradation states of each component (i.e., the ground-truth concepts in our proposed framework). The objective is to train a model to predict the RUL value $\hat{y}$ as a function of intermediate concept representations $\hat{\mathbf{c}}$. In the first step, features are extracted from the input signals by a feature extractor $\phi(\cdot)$. In the second step, concept representations $\hat{\mathbf{c}}$ and their associated activations are derived as detailed in the previous Section~\ref{sec:background}. Finally, a regressor, $f(\cdot)$, computes the RUL value $\hat{y}$. In this work, $\phi$ is modeled by a deep CNN, and $f$ consists of a simple linear layer. The network is trained end-to-end, using the loss function specified in Eq.~\ref{eq:loss}. The first term in the loss is the mean squared error (MSE) between the predicted and true RUL value, adapted for a regression task, rather than the cross-entropy typically used in previous works for classification. The second term is the binary cross-entropy (BCE) between the predicted concept activations and the ground-truth concepts. 

In practical applications, the exact degradation state of an asset is unknown because it is not continuously monitored, making the quantification of the degradation state challenging. However, when an asset is maintained after failure, both the failed component and the failure mode can be identified and documented. Furthermore, ground truth information can also be derived from physics-based performance models. Therefore, we consider information about which components failed and when to be accessible in real-world scenarios, as represented in such a ($\mathbf{x}$, $y$, $\mathbf{c}$) dataset. Furthermore, identifying defective components provides a human-understandable explanation for the decision of the model, thereby enhancing its interpretability and trustworthiness.

\subsection{Completeness of the concepts}

It is important to mention that not all possible degradation modes are always known, and most of the time, labels are not available for many of them. Moreover, in complex systems comprising many sub-components, it may not be obvious which components to include as concepts, and all failure modes cannot be understood. Often, not all the factors that influence the condition monitoring data are known or can be measured \citep{rombach_contrastive_2021}. This is related to the notion of \emph{completeness}: the set of concepts it said to be complete if they are sufficient to accurately predict the target, i.e., they form a complete representation of the task; otherwise, the concepts are incomplete. In real-world industrial scenarios, the concepts are deemed to be incomplete. Nevertheless, as our experiments will demonstrate, concept bottleneck models such as hybrid CBMs and CEMs \citep{zarlenga_concept_2022} can still accurately predict the RUL despite using a very incomplete set of concepts.

\subsection{Test-time intervention strategies}\label{sec:method:intervention}

A key feature of CBMs is the ability to intervene on concepts at test-time \citep{koh_concept_2020,zarlenga_concept_2022}. Concretely, to intervene on concept $i$, one replaces the activations $\hat{\mathbf{c}}_i$ in CBMs or the probabilities $\hat{p}_i$ for CEMs, by the ground-truth activation of the concept $c_i$, determined by a human expert. This results in a new downstream prediction, \emph{corrected} through the intervention. 

The possibility of intervening on the concepts at test-time is particularly invaluable in safety-critical applications such as prognostics. First, an expert is able to simulate multiple scenarios by setting different levels of degradation and predicting the associated RUL values ("what would be the RUL if this degradation mode occurred?"). This \textit{simulatability} is crucial for building trust, and for digital twins in general. Second, whenever the prediction of the model seems wrong because a concept was wrongly activated or not activated, the prediction can be corrected. 

To perform test-time interventions, the user must be able to identify  which concepts are present or absent  in the input. In standard computer vision tasks, it is assumed that the user can visually assess the presence or absence of concepts by examining the input image. In our work, we assume an operator can assess the the true degradation state, characterized by the presence of the component's failure modes, by performing diagnostics or inspecting the asset. For example, in the case of an aircraft engine, inspection can be performed during workshop visits or regular routine inspections. However, since this is an expensive operation, the number of interventions should be kept very limited.
Furthermore, in image classification, samples are typically statistically independent, allowing  interventions to be performed independently on each input. This is not the case for prognostics, where observations are sequential in nature and the state evolves over time. 

\begin{figure}
    \centering
    \includegraphics[width=0.7\linewidth]{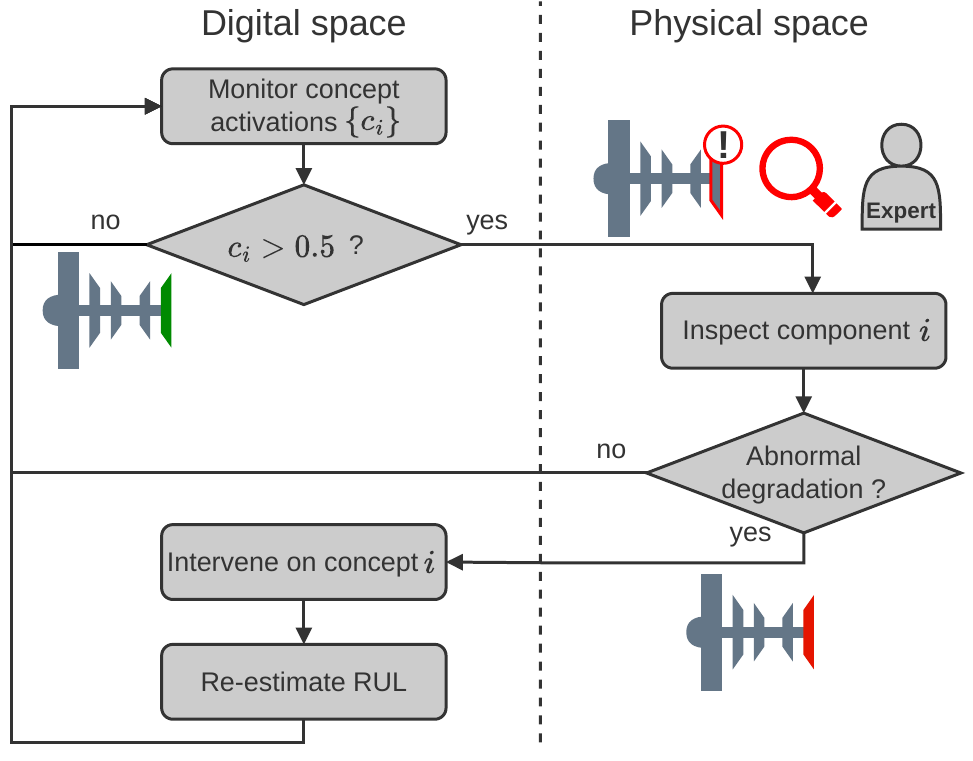}
    \caption{Workflow of a potential concept intervention strategy for prognostics evaluated in this study. After a fault is detected by the model, an operator can inspect the asset and confirm the degradation, and proceed to a new estimation of the RUL given this new knowledge. This re-estimation would be impossible with standard prognostics models.}
    \label{fig:intervention-flow}
\end{figure}

We propose a concept intervention strategy tailored for prognostics, as depicted in Fig.~\ref{fig:intervention-flow}. At each cycle, the concept-based prognostics model predicts the RUL and concept activations for each component of the asset. If a concept's activation  exceeds  the $0.5$ threshold, an expert will inspect the component and determine its actual state. If  the component is indeed degraded, a concept intervention is performed in the model, setting the corresponding concept activation to $1$ for this cycle and for all future cycles, since the component cannot revert  to a non-degraded state on its own. This leads to a revised  estimation of the RUL.

\begin{figure}
    \centering
    \includegraphics[width=0.9\linewidth]{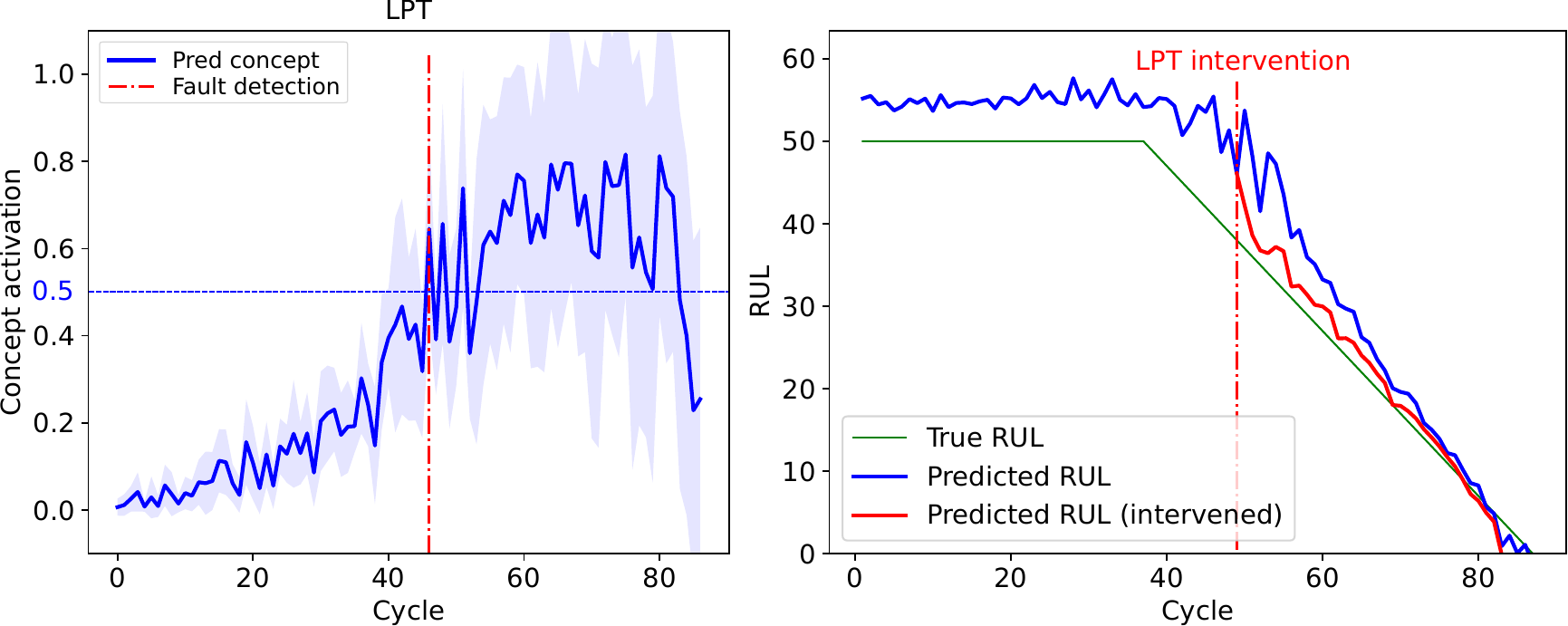}
    \caption{Example of concept intervention on the LPT in DS07 unit 10. A fault is detected on the LPT when the corresponding concept activates over 0.5 during a cycle. After confirming the degradation through an inspection, we intervene on the LPT concept by setting its activation to 1 for the remaining cycles, resulting in a new RUL estimation.}
    \label{fig:intervention-ex}
\end{figure}

An example of an intervention is detailed in Fig.~\ref{fig:intervention-ex}. A fault on the low-pressure turbine (LPT) is detected when the corresponding concept activation exceeds  $0.5$. After confirming the degradation through an inspection or diagnostics, we perform a concept intervention on the LPT concept by setting its activation to 1 for all the remaining cycles, leading to a new RUL estimation. This re-estimated RUL is lower and closer to the true value.

\section{Experiments}\label{sec:experiments}

\subsection{Dataset}

\begin{figure}
    \centering
    \includegraphics[width=0.6\linewidth]{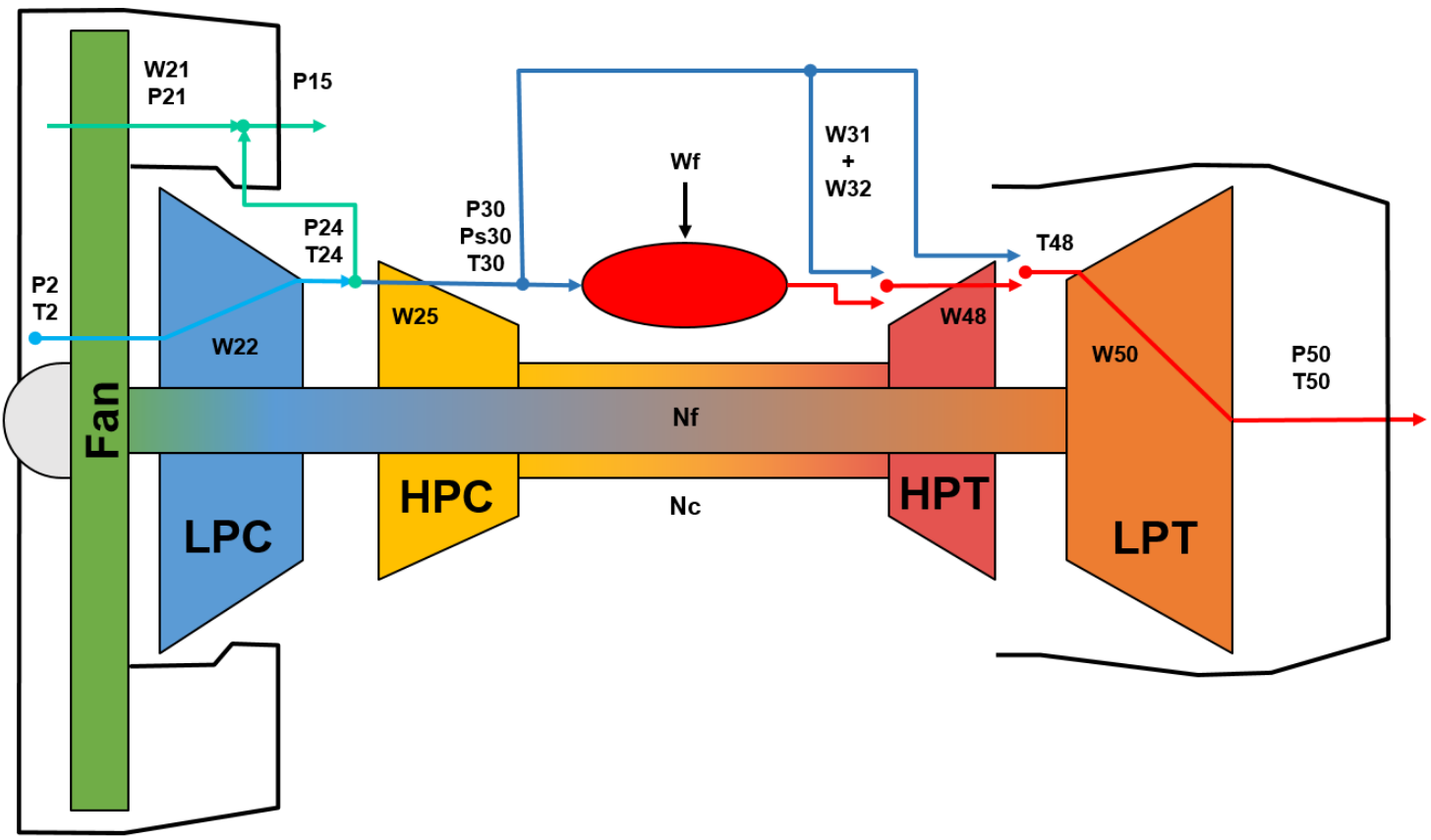}
    \caption{Schematic representation of the CMAPSS model \citep{arias_chao_aircraft_2021}. Air intake enters by the Fan, and a part of the flow passes though the low-pressure compressor (LPC), high-pressure compressor (HPC), combustion chamber, high-pressure turbine (HPT) and low-pressure turbine (LPT).}
    \label{fig:cmapss-schema}
\end{figure}

We conduct experiments on the New Commercial Modular AeroPropulsion System Simulation (N-CMAPSS) dataset \citep{arias_chao_aircraft_2021}, which contains run-to-failure trajectories of a fleet of turbofan engines under real flight conditions generated using the CMAPSS dynamical simulator \citep{saxena_damage_2008}. The architecture of the considered turbofan, with its different stages and measurements, is depicted in Fig.~\ref{fig:cmapss-schema}.

\subsection{Data preparation and preprocessing}

\noindent\textbf{Inputs ($\mathbf{x}$).} Following previous works \citep{arias_chao_fusing_2022}, the input data consists in  14 measurements and 4 operating conditions in a sliding window of size 50 and stride 1, after subsampling by a factor of 10. Zero-padding is used for the very first windows,  resulting in each window covering 500 seconds. We adopt standard scaling to normalize each input variable, based on the overall training data (also called $z$-score normalization), as done in \cite{huang_bidirectional_2019}. This differs from \cite{arias_chao_fusing_2022} who use min-max scaling. We observed that standard scaling leads to faster convergence and slightly higher performance (see~\ref{sec:appendix:scaling}).

\begin{figure}
    \centering
    \includegraphics[width=0.5\linewidth]{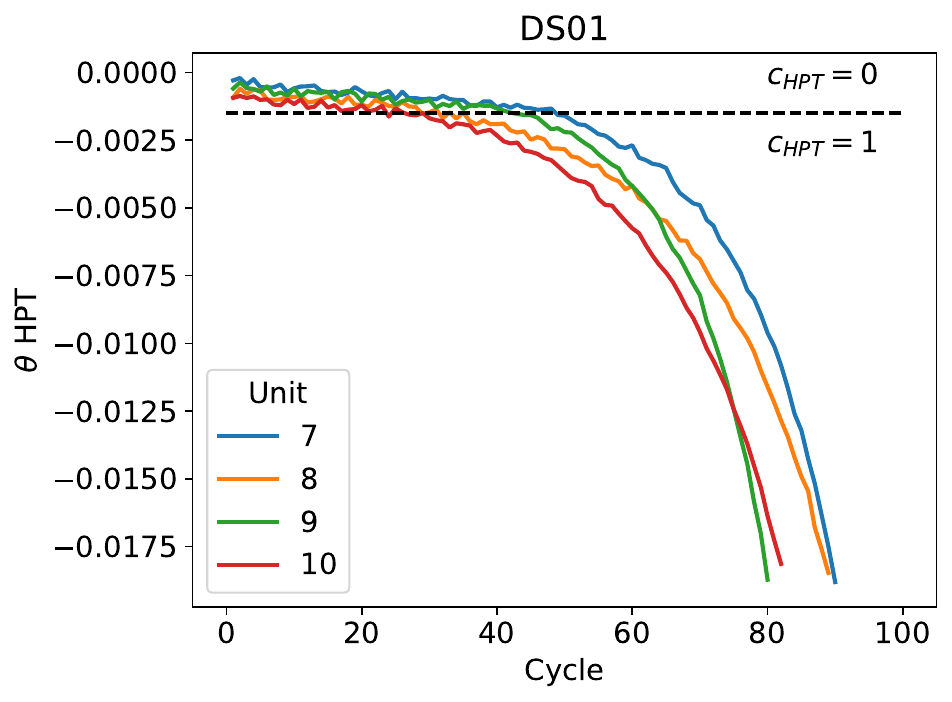}
    \includegraphics[width=0.46\linewidth]{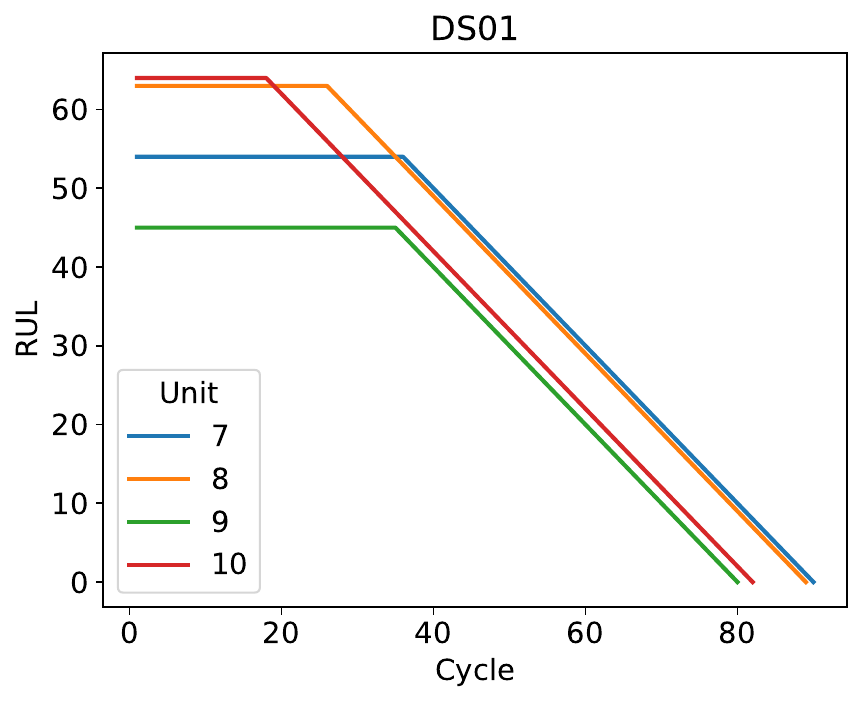}
    \caption{Examples of the degradation trajectories and associated RUL in N-CMAPSS dataset DS01 (test units). The concept label $c$ (here, corresponding to HPT) is obtained by binarizing the degradation level $\theta$ using a fixed threshold of $-0.0015$.}
    \label{fig:degradation}
\end{figure}

\noindent\textbf{Target ($y$).} The true RUL target values are labeled following a piece-wise linear RUL target function \citep{shi_dual-lstm_2021}. Before the beginning of abnormal degradation (i.e., the change of health state \texttt{hs} in the N-CMAPSS dataset), the RUL is defined as constant, indicating  that the unit experiences only a negligible amount of degradation before this point. During the abnormal degradation phase, the RUL linearly decreases until reaching zero (see Fig.~\ref{fig:degradation}).

\noindent\textbf{Concepts ($\mathbf{c}$).} To obtain a set of binary concepts for each input window, we binarize the degradation parameters $\theta$ for each component of the engine. In N-CMAPSS, each component can experience two types of degradation: the efficiency modifier (\texttt{eff\_mod}) and the flow modifier (\texttt{flow\_mod}). We aggregate both degradations into a single degradation level per component by taking the minimum, e.g., for the HPT: $\theta_{\text{HPT}} = \min (\theta_{\text{HPT\_eff\_mod}}, \theta_{\text{HPT\_flow\_mod}}$. We observed in our experiments that keeping both degradation parameters distinct does not change the results and would only harm the clarity of this paper. Then, to obtain a binary concept, we binarize each degradation parameter using a threshold fixed at $\tau = -0.0015$, e.g., the corresponding HPT concept is defined as $c_{\text{HPT}} = \mathbbm{1}_{\theta_{\text{HPT}} \leq \tau}$. This allows us to obtain one binary attribute per engine component; it is, however, different from the fault onset. 

Examples of degradation trajectories and associated concepts and RUL target values in the dataset DS01 are illustrated in Fig.~\ref{fig:degradation}. The input target RUL values are scaled by a fixed factor of 100. When reporting the performance of RUL prediction, the values are unscaled.



%
\subsection{Evaluation metrics}

\noindent\textbf{RUL prediction.} We adopt common evaluation metrics in prognostics \citep{saxena_damage_2008,arias_chao_aircraft_2021} to evaluate the performance of the RUL prediction task. First, we evaluate the Root Mean Square Error (RMSE) per cycle. As our model outputs a RUL value at each input window, we average the predictions within each cycle $q$ to obtain $\bar{y}^{(q)}$ and compute the error as $\text{RMSE} = \sqrt{\frac{1}{n_{\text{cycles}}}\sum_{q=1}^{n_{\text{cycles}}} \left(\bar{y}^{(q)} - y^{(q)}\right)^2}$, where $n_{\text{cycles}}$ is the total number of cycles. We also evaluate NASA's scoring function defined as $\text{NASA} = \frac{1}{n_{\text{cycles}}} \sum_{q=1}^{n_{\text{cycles}}} \exp \left( \alpha |\bar{y}^{(q)} - y^{(q)}|\right) - 1$, where $\alpha = \frac{1}{13}$ if the RUL is under-estimated and $\alpha = \frac{1}{10}$ if it is over-estimated. As a result, this scoring penalizes over-estimation more than under-estimation.

\noindent\textbf{Concept classification.} To evaluate the quality of concept classification, we use multi-label classification accuracy. The accuracy for concept $j$ is defined as $\text{Acc}(j) = \frac{1}{n} \sum_{i=1}^n \mathbbm{1}(\hat{c}_j^{(i)} = c_j^{(i)})$. The overall accuracy is obtained by macro-averaging across all concepts: $\text{Acc} = \frac{1}{k}\sum_{j=1}^k \text{Acc}(j)$.

\noindent\textbf{Concept alignment.} We evaluate the Concept Alignment Score (CAS) \citep{zarlenga_concept_2022}  to assess how well the concept activations (for CBMs), concept embeddings (for CEM), or latent code (for CNN and CNN+CLS) are aligned with the concepts. 

\noindent\textbf{Fault detection.} When assessing the fault detection performance, we use the area under the curve of the receiver operating characteristic (AUC-ROC) of the predicted fault score compared with the ground-truth health state at each cycle. The health state  is equal to zero before the fault onset and one after the onset, as represented by the health state (\texttt{hs}) variable in the N-CMAPSS dataset.

\subsection{Experimental setting}

\begin{table}
    \setlength{\aboverulesep}{0pt}
    \setlength{\belowrulesep}{0pt}
    \renewcommand{\arraystretch}{1.2}
    \centering
    \begin{tabular}{c|cccc|cc}
        \toprule
        \multirow{2}*{Dataset} & \multirow{2}*{\includegraphics[height=2.5em]{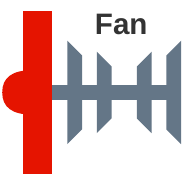}} & \multirow{2}*{\includegraphics[height=2.5em]{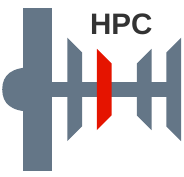}} & \multirow{2}*{\includegraphics[height=2.5em]{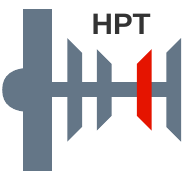}} & \multirow{2}*{\includegraphics[height=2.5em]{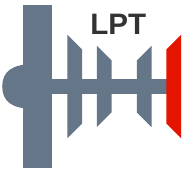}} & 
        \multirow{2}*{Scenario 1} & \multirow{2}*{Scenario 2} \\
        & & & & & & \\
        \midrule
         DS01 & & & \cmark & & \cmark & \cmark\\
         DS03 & & & \cmark & \cmark &  & \cmark \\
         DS04 & \cmark & & & & \cmark & \\
         DS05 & & \cmark & & & \cmark & \\
         DS07 & & & & \cmark & \cmark & \cmark \\
         \bottomrule
    \end{tabular}
    \caption{Overview of the component degradations present in each dataset of N-CMAPSS and the two scenarios studied in this work. Scenario 1 contains four single fault types, while Scenario 2 tackles two single and a combined fault.}
    \label{tab:faults-scenarios}
\end{table}

We evaluate two scenarios in this work -- one containing several single-component faults, and another containing a combined fault on two components, -- summarized in Tab.~\ref{tab:faults-scenarios}.

\noindent\textbf{Scenario 1: Fan, HPC, HPT and LPT degradation}. In this scenario, we combine four datasets (DS01, D04, DS05 and DS07) representing four single-component degradation modes corresponding to the Fan, HPC, HPT and LPT. Hence, four associated concepts are used unless specified otherwise. 

\noindent\textbf{Scenario 2: HPT, LPT and combined HPT+LPT degradation}. This scenario aims to tackle the case of a combined degradation on two components, namely HPT and LPT. Therefore, we use three datasets (DS01, DS03 and DS07) containing units experiencing  single-mode degradations on HPT and LPT, as well as the combined HPT+LPT degradation (in DS03).

Note that the LPC is not included in our study, as there is no dataset in N-CMAPSS with a single LPC fault. For each scenario, we include  units 1, 2, 3, 4, 5, 6 of each dataset in our training set and units 7, 8, 9, 10 of each dataset in our test set.


%
\subsection{Compared methods}

\begin{table}
    \centering
    \resizebox{\linewidth}{!}{
    \begin{tabular}{lccccc}
        \toprule
        \multirow{2}*{Method} & \multicolumn{2}{c}{Prediction} & \multirow{2}*{Interpretable} & Bottleneck & Concept \\
        & RUL & Failure mode & & capacity & activation\\
        \midrule
        CNN & \cmark & \xmark & no & - &  - \\
        CNN+CLS & \cmark & \cmark & no & - & sigmoid \\
        Boolean CBM & \cmark & \cmark & yes & $k$ & hard threshold \\
        Fuzzy CBM & \cmark & \cmark & yes & $k$ & sigmoid \\
        Hybrid CBM & \cmark & \cmark & partially & $k + e$ & sigmoid \\
        CEM & \cmark & \cmark & yes & $k \cdot m$ & sigmoid \\
         \bottomrule
    \end{tabular}
    }
    \caption{Summary of the compared methods. The bottleneck capacity is the number of dimensions of the intermediate bottleneck, where $k$ is the number of concepts, $m$ is the concept embedding dimension in CEM, and $e$ is the extra capacity in hybrid CBM. In all experiments, we set $e$ in order to have $k + e = k \cdot m$.}
    \label{tab:methods-summary}
\end{table}

\begin{figure}
    \centering
    \begin{subfigure}[T]{0.49\textwidth}
         \centering
         \includegraphics[height=13em]{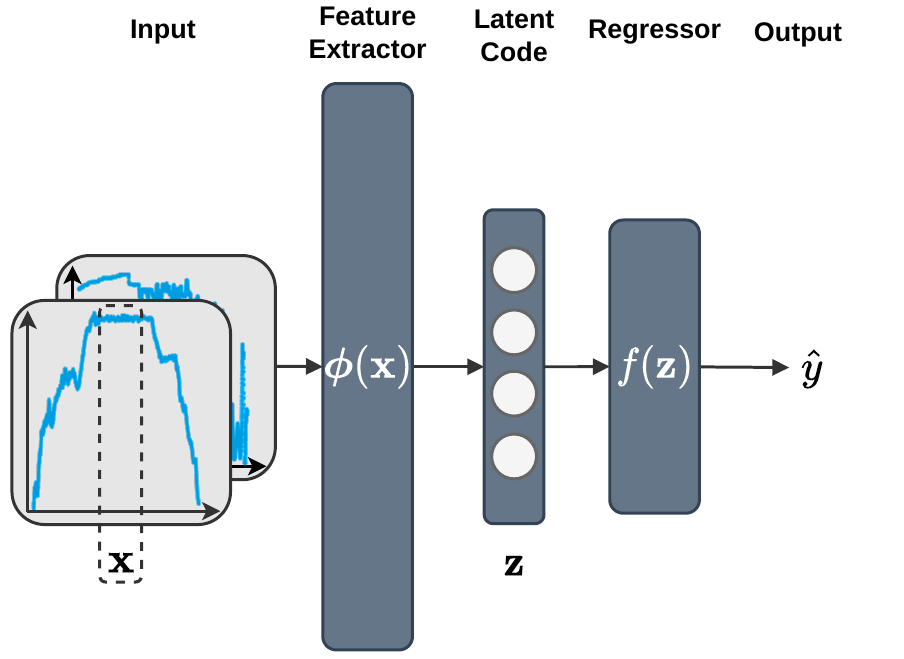}
         \caption{CNN}
         \label{fig:architecture-cnn}
     \end{subfigure}
    \begin{subfigure}[T]{0.5\textwidth}
         \centering
         \includegraphics[height=13em]{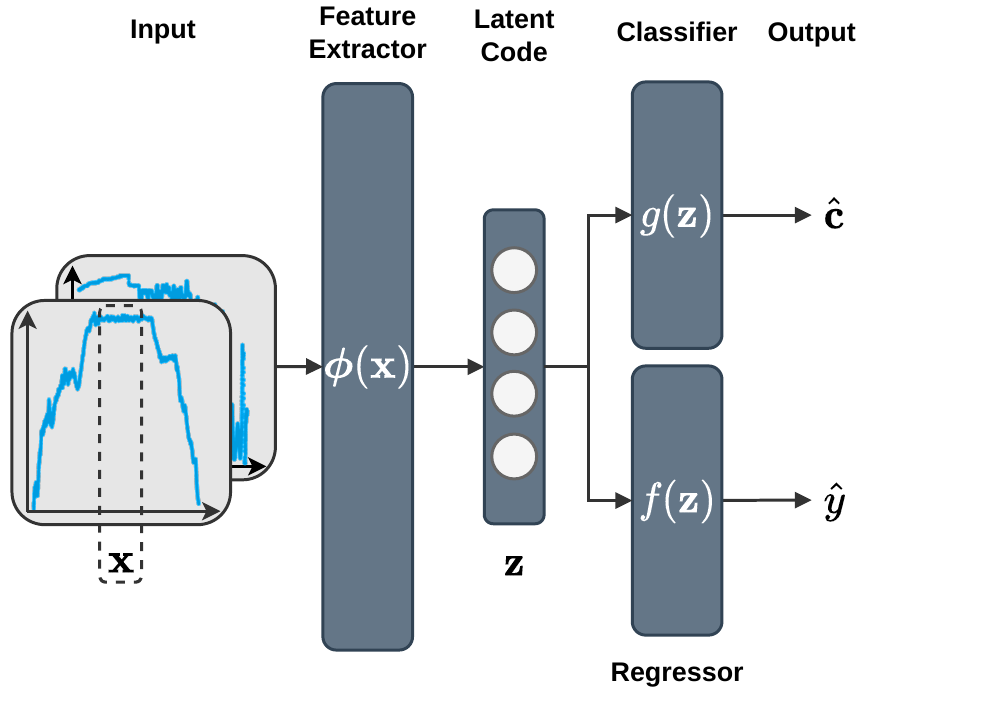}
         \caption{CNN+CLS}
         \label{fig:architecture-cnn-cls}
     \end{subfigure}
    \caption{Non-interpretable architectures for prognostics. The convolutional neural network (CNN) regression model directly predicts the RUL from an input window, while CNN+CLS is a multi-task model with an additional classification head to classify failure modes.}
    \label{fig:architectures-cnn}
\end{figure}

We evaluate and compare several non-interpretable and concept-based methods for RUL prediction, summarized in Tab.~\ref{tab:methods-summary}. All method comprise a 1D-convolutional feature extractor $\phi$ whose architecture is similar to the one used by \cite{arias_chao_fusing_2022}.

\textbf{CNN.} The baseline method is a convolutional neural network (CNN) that only predicts the RUL. It is composed of the feature extractor $\phi$ followed by a regression head $f$, parameterized by a single linear layer. For a given input $\mathbf{x}$, the predicted RUL is computed as $\hat{y} = f(\phi(\mathbf{x}))$. This model is considered as non-interpretable because it does not provide any explanations on how the RUL value is predicted. The CNN is trained with the MSE loss between the true and predicted RUL:

\begin{equation}
    \mathcal{L}_{\text{CNN}} = \text{MSE}(y, \hat{y}) = \frac{1}{n}\sum_{i=1}^n \left(y^{(i)} - \hat{y}^{(i)}\right)^2.
\end{equation}

\textbf{CNN+CLS.} The second considered baseline is a multi-task architecture that simultaneously predicts both the RUL and the degradation modes (i.e., the concepts). This is achieved by adding a classification head $g$ (parameterized by a linear layer with sigmoid activation), taking as input the same intermediate features as the regression head and outputting one probability for each failure mode. Concretely, for a given input $\mathbf{x}$, the predicted RUL and class probabilities are $\hat{y} = f(\phi(\mathbf{x}))$ and $\mathbf{\hat{c}} = g(\phi(\mathbf{x}))$, respectively. We consider this task as multi-label binary classification, similarly to the concept classification in CBMs. Hence, CNN+CLS is trained with a hybrid loss composed of a weighted sum of the MSE on the RUL and the binary cross-entropy (BCE) between the true and predicted class probabilities:

\begin{align}
    \mathcal{L}_{\text{CNN+CLS}} &= \text{MSE}(y, \hat{y}) + \lambda \cdot \text{BCE}(\mathbf{c}, \mathbf{\hat{c}}) \\
    &= \frac{1}{n}\sum_{i=1}^n \left(y^{(i)} - \hat{y}^{(i)}\right)^2 + \frac{\lambda}{n}\sum_{i=1}^n \sum_{j=1}^k c^{(i)}_j \log \hat{c}^{(i)}_j + (1-c^{(i)}_j) \log (1-\hat{c}^{(i)}_j) \nonumber
\end{align}

where $\lambda$ is a weighting coefficient. Both architectures are represented on Fig.~\ref{fig:architectures-cnn}.

\textbf{Boolean/Fuzzy CBM, Hybrid CBM and CEM} as described in Section~\ref{sec:background} and illustrated on Fig.~\ref{fig:architectures-concept}.

\subsection{Training parameters}

All models are trained end-to-end from scratch using the Adam optimizer, a learning rate of $10^{-3}$, and a batch size of 256. In CNN+CLS, CBMs and CEM, the weighting coefficient $\lambda$ is set to $0.1$  to scale the loss terms to the same order of magnitude. We keep  this parameter fixed across all models and scenarios without further tuning. In CEM, the embedding dimension is set to $m = 16$ for each concept, and in hybrid CBMs, the additional capacity is adjusted to match the capacity of the CEM.

The feature extractor $\phi$ is a CNN composed of two $3 \times 3\times 20$ 1D-conv layers followed by two $3 \times 3\times 10$ 1D-conv layers, and finally a linear layer to map to the latent space. ReLU activations are applied between each layer. The latent code dimension for the CNN, CNN+CLS and CEM is 256. The regressor $f$ and classifier $g$ are linear layers mapping to one scalar output for the RUL regression (without activation), and $k$ outputs with sigmoid activation for the classifier in the CNN+CLS model. 

\section{Results}\label{sec:results}

This section presents  the results of our experiments. We  evaluate and compare the different models based on RUL estimation performance, concept accuracy, concept alignment, and fault detection performance. Additionally, we assess  the effectiveness of test-time concept interventions and perform an ablation study on the number of concepts used in training. Finally, we visualize the embedding space of a trained CEM model to qualitatively assess the structure of the learned concept embeddings.

\subsection{RUL estimation performance}
%

\begin{table}
    \setlength{\aboverulesep}{0pt}
    \setlength{\belowrulesep}{0pt}
    \centering
    \begin{tabular}{lcc|cc}
        \toprule
         & \multicolumn{2}{c|}{Scenario 1} & \multicolumn{2}{c}{Scenario 2} \\
        Method & RMSE & NASA & RMSE & NASA \\ 
        \midrule
        CNN	& 6.79 & 0.851 & 6.41 & \textbf{0.703} \\
        CNN+CLS	& 6.27 & 0.755 & \textbf{6.23} & 0.724 \\
        \midrule
        Boolean CBM	& 12.50 & 2.334  & 11.20 & 1.896 \\
        Fuzzy CBM & 8.08 & 1.168 & 8.13 & 1.009 \\
        Hybrid CBM & 6.65 & 0.827 & 6.46 & 0.733 \\
        CEM & \textbf{6.15} & \textbf{0.752} & 6.37 & 0.728 \\
        \bottomrule
    \end{tabular}
    \caption{RUL prediction performance (RMSE and NASA score in cycles, lower is better) for different methods on Scenario 1 and Scenario 2. Average across all test units.}
    \label{tab:results-rul-1457}
\end{table}

The average performance for RUL prediction across all datasets in Scenario 1 is reported in Tab~\ref{tab:results-rul-1457}. Among the compared methods, the CEM achieved the lowest RMSE with an average of 6.15 cycles, which is slightly better than the non-interpretable models (6.79 for CNN and 6.27 for CNN+CLS). The Boolean and fuzzy CBMs perform poorly due to their limited expressive power, while the hybrid CBM is also competitive. Interestingly, adding a classification head improves RUL prediction performance, showing that multi-task learning allows the CNN+CLS to generalize better than the CNN alone. Conclusions are similar when examining the NASA score.

\begin{figure}
    \centering
    \begin{subfigure}{0.3\linewidth}
        \includegraphics[width=\linewidth]{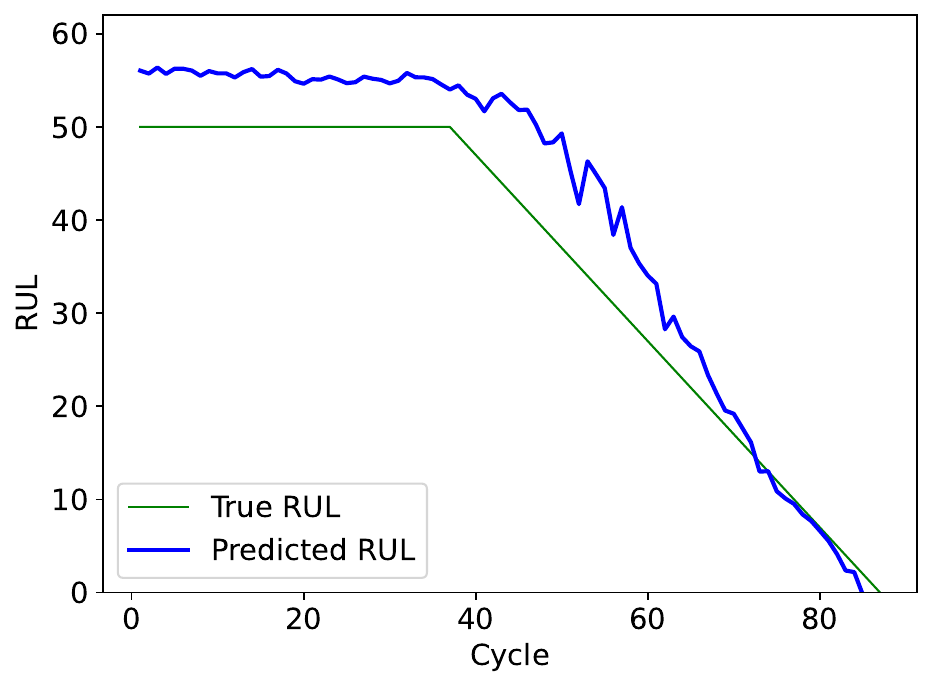}
        \caption{CNN}
        \label{fig:rul:cnn}
    \end{subfigure}
    \begin{subfigure}{0.3\linewidth}
        \includegraphics[width=\linewidth]{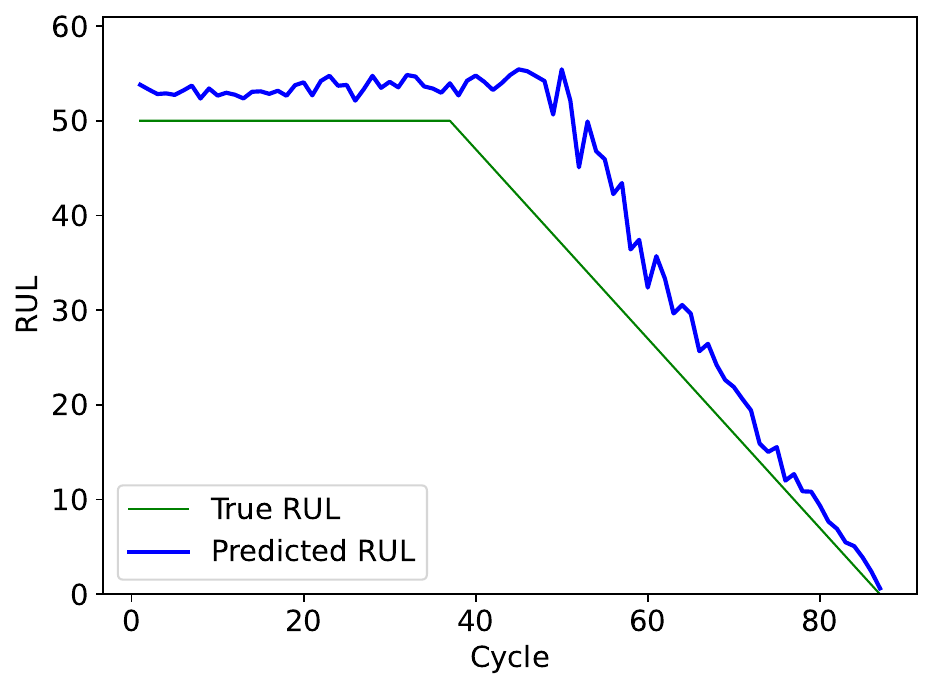}
        \caption{CNN+CLS}
        \label{fig:rul:cnncls}
    \end{subfigure}
    \begin{subfigure}{0.3\linewidth}
        \includegraphics[width=\linewidth]{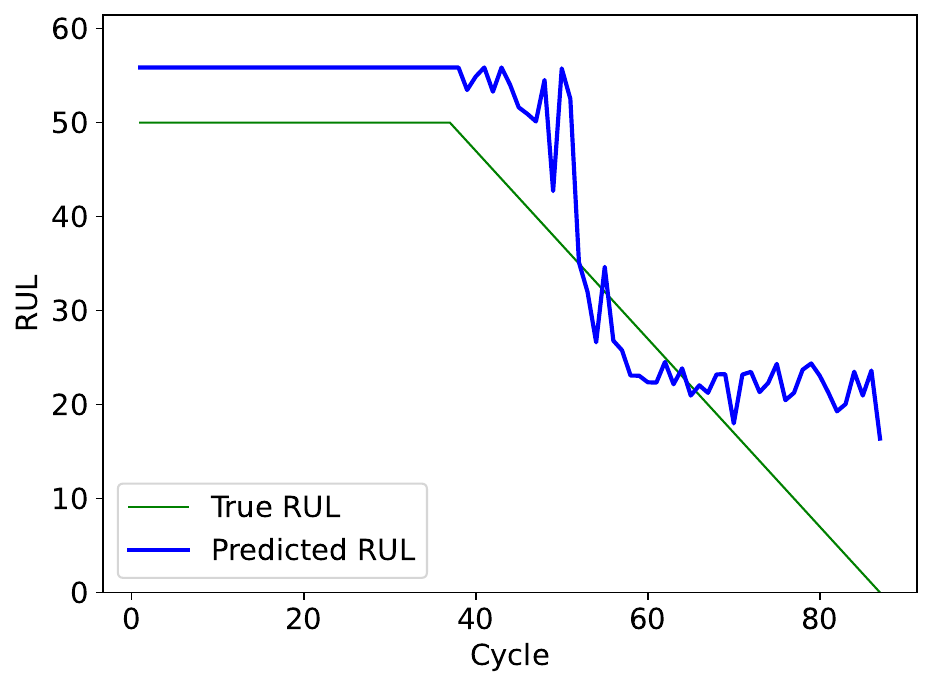}
        \caption{Boolean CBM}
        \label{fig:rul:cbmbool}
    \end{subfigure}
    \begin{subfigure}{0.3\linewidth}
        \includegraphics[width=\linewidth]{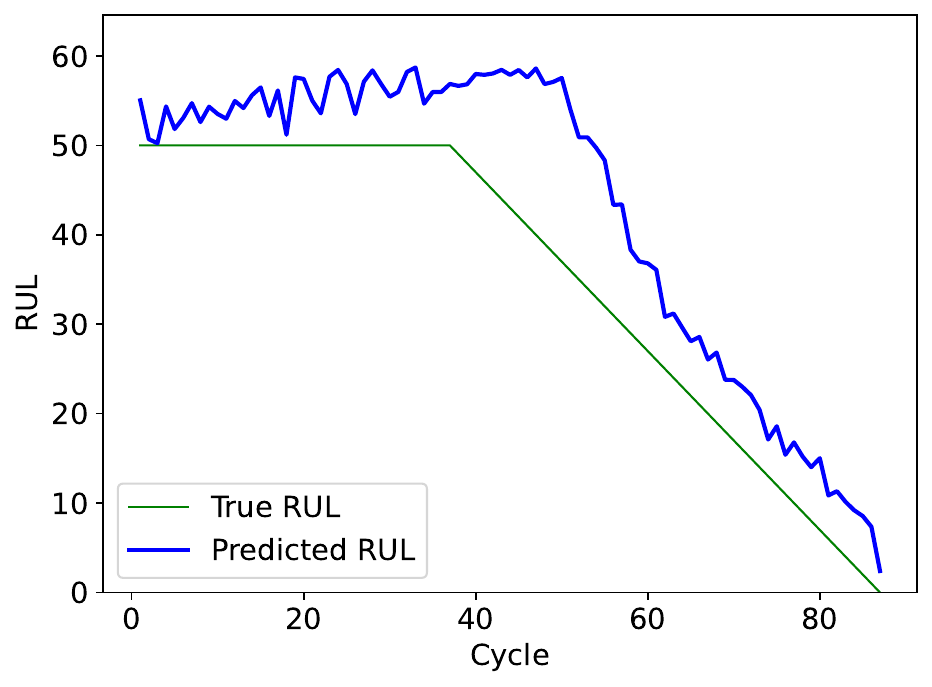}
        \caption{Fuzzy CBM}
        \label{fig:rul:cbm}
    \end{subfigure}
    \begin{subfigure}{0.3\linewidth}
        \includegraphics[width=\linewidth]{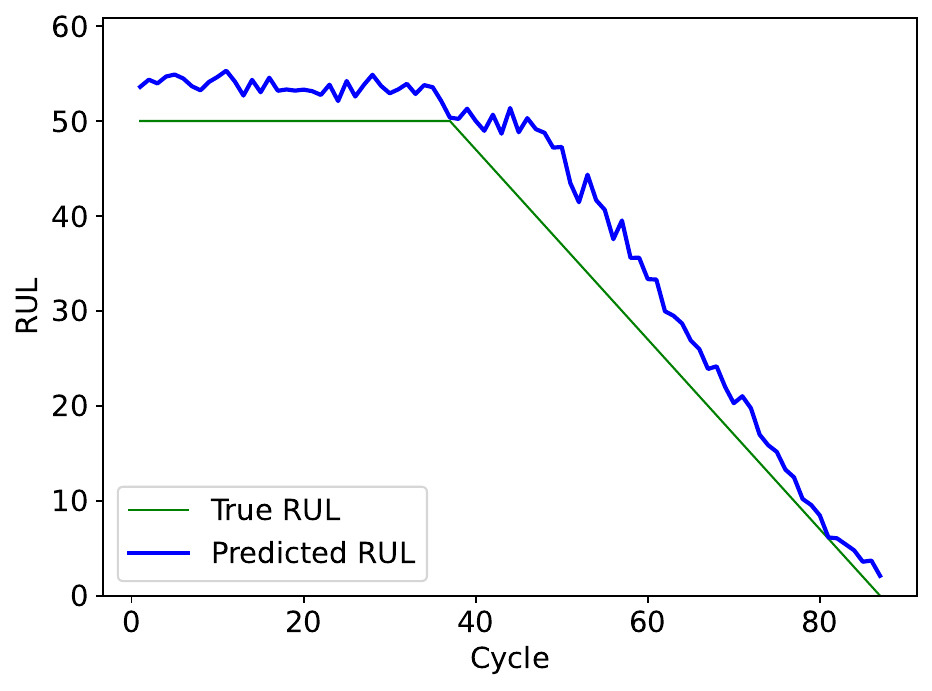}
        \caption{Hybrid CBM}
        \label{fig:rul:cbmhybrid}
    \end{subfigure}
    \begin{subfigure}{0.3\linewidth}
        \includegraphics[width=\linewidth]{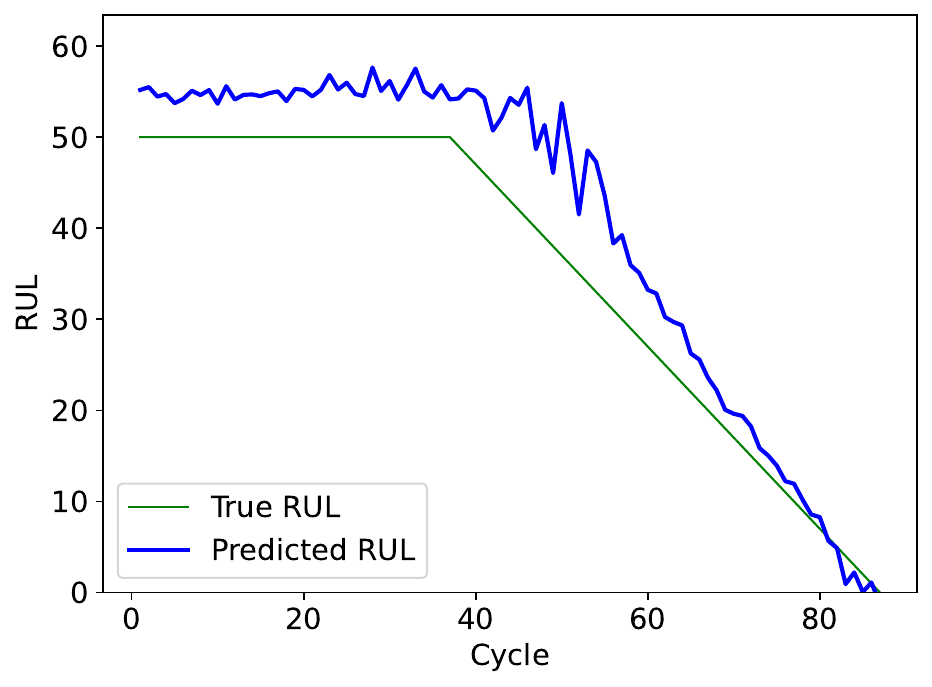}
        \caption{CEM}
        \label{fig:rul:cem}
    \end{subfigure}
    \caption{RUL trajectories for unit 10 in DS07 predicted by different methods.}
    \label{fig:rul}
\end{figure}

Fig.~\ref{fig:rul} displays an example of the predicted RUL trajectory for unit 10 in DS07 for each of the compared methods. The Boolean CBM only handles Boolean values in its bottleneck, preventing it from learning the mapping to the RUL value. The non-interpretable CNN and CNN+CLS, as well as the interpretable hybrid CBM and CEM, exhibit the most accurate RUL predictions.


In the second scenario, which contains a combined degradation, the CEM performs best among the interpretable methods but slightly worse than CNN+CLS. The CEM and the hybrid CBM remain competitive with the CNN and CNN+CLS baselines, showing that in this more complex scenario, there is still no significant trade-off between performance and interpretability. More detailed results for each dataset and unit are reported in \ref{sec:appendix:results}.

\subsection{Concept accuracy}

\vspace{0.25cm}
\begin{minipage}{0.6\linewidth}
        \setlength{\tabcolsep}{2pt}
        \centering
        \resizebox{\linewidth}{!}{
        \begin{tabular}{lccccc}
            \toprule
             \raisebox{1em}{Method} & \includegraphics[height=3em]{fan.pdf} & \includegraphics[height=3em]{HPC.pdf} & \includegraphics[height=3em]{HPT.pdf} & \includegraphics[height=3em]{LPT.pdf} & \raisebox{1em}{Average} \\
            \midrule
            CNN+CLS & \textbf{95.30} & \textbf{98.02} & 97.13 & 93.67 & 96.03 \\
            \midrule
            Boolean CBM & 93.20 & 97.95 & 97.06 & 94.55 & 95.69 \\
            Fuzzy CBM & 92.42 & 95.54 & 93.13 & 90.09 & 92.79 \\
            Hybrid CBM & 91.77 & 94.74 & 95.01 & 90.22 &  92.94 \\
            CEM & 95.15 & 96.46 & \textbf{98.18} & \textbf{95.12} & \textbf{96.23} \\
             \bottomrule
        \end{tabular}
        }
        \captionof{table}{Concept accuracy (\%, higher is better) for different methods in Scenario 1. Average across all test units in DS\{01,04,05,07\}.}
        \label{tab:results-acc-1457}
\end{minipage}
\begin{minipage}{0.39\linewidth}
        \centering
        \includegraphics[width=\linewidth]{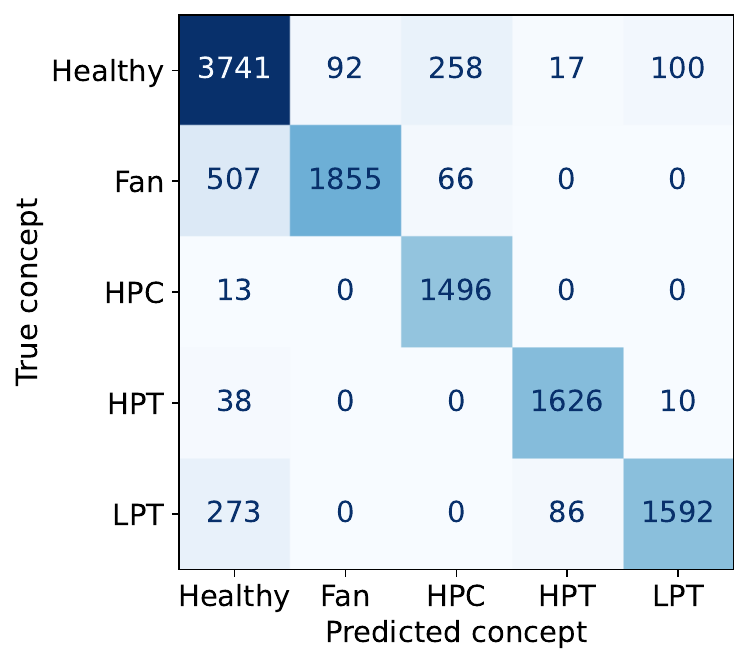}
        \captionof{figure}{CEM confusion matrix.}
        \label{fig:confusion-cem-1457}
\end{minipage}
\vspace{0.5cm}

In Scenario 1, all methods achieve very high concept accuracy, with the CEM achieving the highest accuracy of 96.23\% (see Tab.~\ref{tab:results-acc-1457}). We further analyze the concept classification of CEM in the confusion matrix Fig.~\ref{fig:confusion-cem-1457}, where we defined a healthy class corresponding to the absence of any active concept. Most errors come from confusing between the different degradation modes and the healthy class, which is expected when the level of degradation is still very close to the healthy state.

\vspace{0.25cm}
\begin{minipage}{0.6\linewidth}
    \setlength{\tabcolsep}{2pt}
    \centering
    {\scriptsize
    \begin{tabular}{lccc}
        \toprule
         \raisebox{1em}{Method} & \includegraphics[height=3em]{HPT.pdf} & \includegraphics[height=3em]{LPT.pdf} & \raisebox{1em}{Average} \\
         \midrule
         CNN+CLS & 84.17 & 85.93 & 85.05 \\
         \midrule
         Boolean CBM & 81.48 & 74.51 & 78.00 \\
         Fuzzy CBM & 84.16 & 82.02 & 83.09 \\
         Hybrid CBM & \textbf{84.22} & \textbf{86.27} & \textbf{85.25} \\
         CEM & 83.87 & 83.64 & 83.75 \\
         \bottomrule
    \end{tabular}
    }
    \captionof{table}{Concept accuracy (\%, higher is better) for different methods in scenario 2. Average across all test units in DS\{01,03,07\}.}
    \label{tab:results-acc-137}
\end{minipage}
\begin{minipage}{0.39\linewidth}
        \centering
        \includegraphics[width=\linewidth]{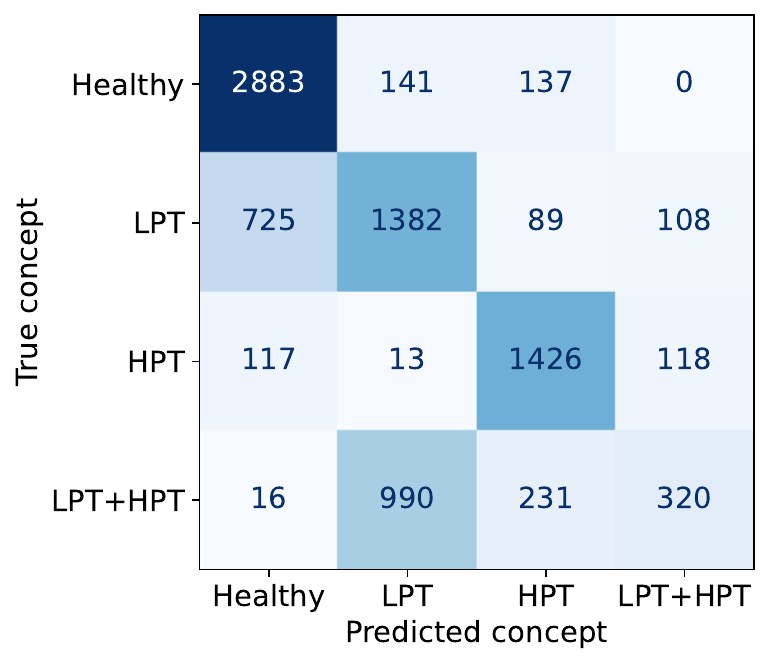}
        \captionof{figure}{CEM confusion matrix.}
        \label{fig:confusion-cem-137}
\end{minipage}
\vspace{0.5cm}

\begin{figure}
    \begin{subfigure}{0.36\linewidth}
        \centering
        \includegraphics[width=\linewidth]{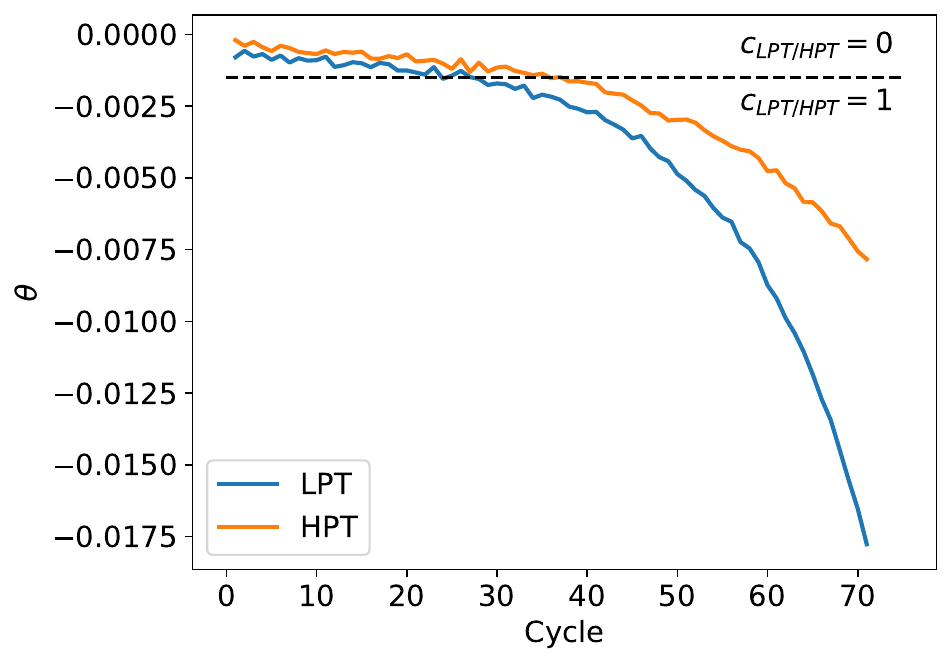}
        \caption{Degradation parameters.}
        \label{fig:combined-fault:theta}
    \end{subfigure}
    \begin{subfigure}{0.65\linewidth}
        \centering
        \includegraphics[width=\linewidth]{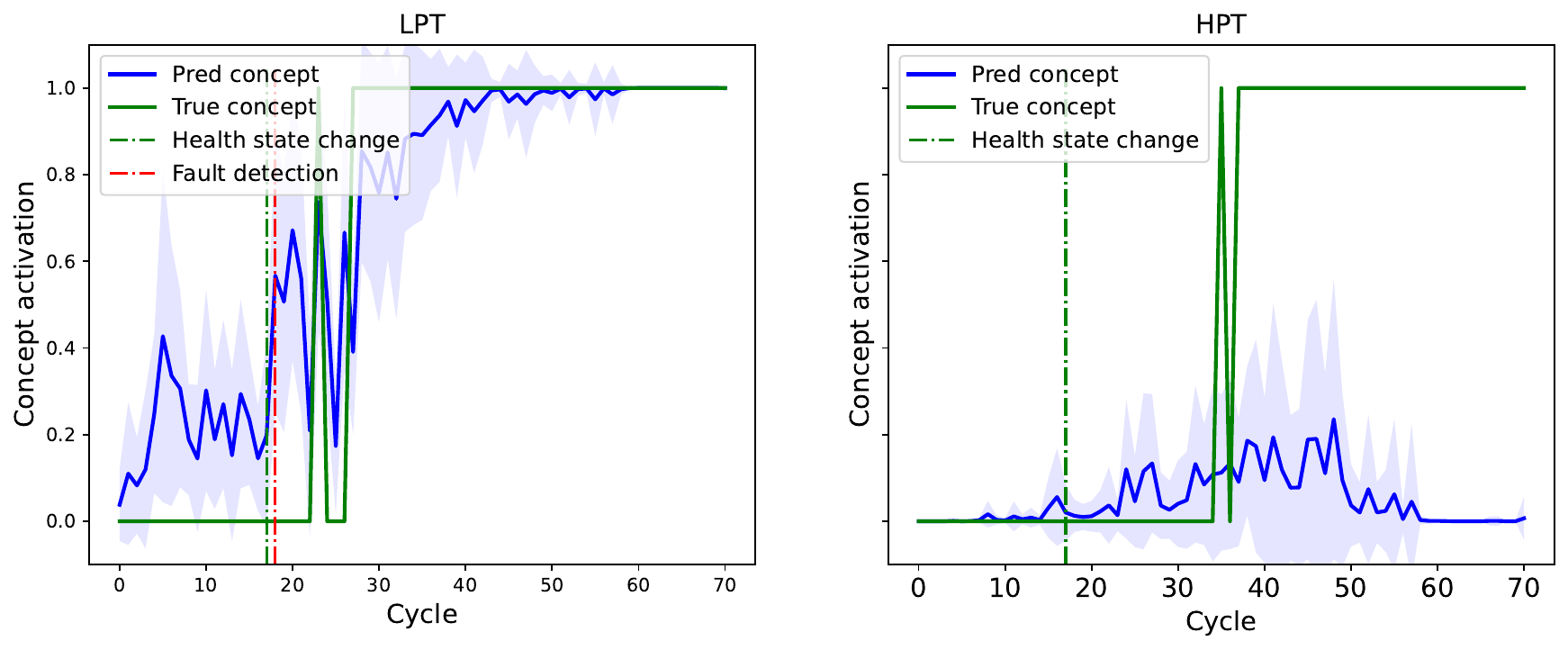}
        \caption{Concept activations in a CEM trained in Scenario 2.}
        \label{fig:combined-fault:concept}
    \end{subfigure}
    \caption{In the case of combined faults, (here, LPT+HPT fault in unit 8 of DS03), only the concept associated with the dominant degradation mode is being activated (here, LPT), although both binary concepts are active in the ground-truth.}
    \label{fig:combined-fault}
\end{figure}

In Scenario 2, the concept accuracy is substantially lower, at around 85\% (see Tab.~\ref{tab:results-acc-137}). To understand this result, we build a confusion matrix by defining an additional class LPT+HPT corresponding to the combination of both LPT and HPT degradations active at the same time. Note that the combined degradation does not correspond to a concept by itself and  is only defined to obtain mutually exclusive (not multi-label) classes in the confusion matrix shown Fig.~\ref{fig:confusion-cem-137}. The main observation is that most combined LPT+HPT faults are classified as either LPT or HPT faults. Our interpretation is that only one concept will activate, depending on which is the most dominant degradation among both. In the combined degradation trajectories of DS03, for most units, one degradation mode dominates the other, with a lower value of the degradation parameter. As illustrated in Fig.~\ref{fig:combined-fault}, LPT degradation is more severe than the HPT one. While both ground-truth concepts are active, the model only activates the most severe one. 

Moreover, the different models also confuse between the single HPT and LPT faults, which are harder to distinguish since they have similar signatures in the input signals.

\subsection{Concept alignment}

\vspace{0.25cm}
\begin{minipage}[t]{0.6\linewidth}
    \setlength{\tabcolsep}{2pt}
    \centering
    {\scriptsize
    \begin{tabular}{lccccc}
        \toprule
         \raisebox{1em}{Method} & \includegraphics[height=3em]{fan.pdf} & \includegraphics[height=3em]{HPC.pdf} & \includegraphics[height=3em]{HPT.pdf} & \includegraphics[height=3em]{LPT.pdf} & \raisebox{1em}{Average} \\
        \midrule
        CNN & 62.45 & 58.68 & 59.20 & 58.03 & 59.59 \\
        CNN+CLS & \textbf{85.67} & 88.09 & 86.75 & 82.55 & 85.77 \\
        \midrule
        Boolean CBM & 77.95 & \textbf{92.37} & 89.25 & 81.12 & 85.17 \\
        Fuzzy CBM & 72.87 & 80.05 & 77.63 & 72.75  & 75.82 \\
        Hybrid CBM & 82.99 & 88.25 & 83.61 & 77.78 & 83.16 \\
        CEM & 85.51 & 91.39 & \textbf{90.98} & \textbf{85.35} & \textbf{88.31} \\
         \bottomrule
    \end{tabular}
    }
    \captionof{table}{Concept alignment score (\%, higher is better) for different methods on Scenario 1. Average across all test units in DS\{01,04,05,07\}.}
    \label{tab:results-cas-1457}
\end{minipage}
\quad
\begin{minipage}[t]{0.4\linewidth}
    \setlength{\tabcolsep}{2pt}
    \centering
    {\scriptsize
    \begin{tabular}{lccc}
        \toprule
         \raisebox{1em}{Method} & \includegraphics[height=3em]{HPT.pdf} & \includegraphics[height=3em]{LPT.pdf} & \raisebox{1em}{Average} \\
        \midrule
        CNN & 65.11 & 61.55 & 63.33 \\
        CNN+CLS & 76.46 & \textbf{91.38} & \textbf{83.92} \\
        \midrule
        Boolean CBM & 49.46 & 57.54 & 53.50 \\
        Fuzzy CBM & 57.04 & 64.54 & 60.79 \\
        Hybrid CBM & 75.94 & 85.56 & 80.75 \\
        CEM & \textbf{79.27} & 86.36 & 82.82 \\
         \bottomrule
    \end{tabular}
    }
    \captionof{table}{Concept alignment score (\%, higher is better) for different methods on Scenario 2. Average across all test units in DS\{01,03,07\}.}
    \label{tab:results-cas-137}
\end{minipage}
\vspace{0.5cm}

\subsection{Fault detection}

Prognostics models integrating fault classification are inherently able to perform fault detection by considering the highest fault probability as a fault detection score. This includes the CNN with the classification head (CNN+CLS), the CBMs and the CEM. As explained in the previous section, we consider a fault is detected when a concept has an activation higher than 0.5, as visualized, for example, in Fig.~\ref{fig:combined-fault:concept}.
\begin{table}
    \setlength{\aboverulesep}{0pt}
    \setlength{\belowrulesep}{0pt}
    \centering
    \begin{tabular}{lc|c}
        \toprule
        & Scenario 1 & Scenario 2 \\
        Method & AUC & AUC \\ 
        \midrule
        CNN+CLS	& 93.13 & 94.72 \\
        \midrule
        Boolean CBM	& 92.84 & 95.79 \\
        Fuzzy CBM & 86.46 & 95.12 \\
        Hybrid CBM & 90.11 & \textbf{96.31} \\
        CEM & \textbf{93.49} & 94.53 \\
        \bottomrule
    \end{tabular}
    \caption{Fault detection performance (AUC in \%, higher is better) for different methods on Scenario 1 and Scenario 2. Average across all test units.}
    \label{tab:results-fault}
\end{table}
The results reported in Table~\ref{tab:results-fault} suggest that all methods have high fault detection accuracy, meaning that the activation of the concepts highly correspond to the true health state of the system.

\subsection{Test-time concept interventions}

We evaluate the test-time concept intervention strategy proposed in Section~\ref{sec:method:intervention}. We take the average activation of the concepts during each cycle as the concept activation for that cycle.
\begin{figure}
    \centering
    \begin{subfigure}{0.45\linewidth}
        \includegraphics[width=\linewidth]{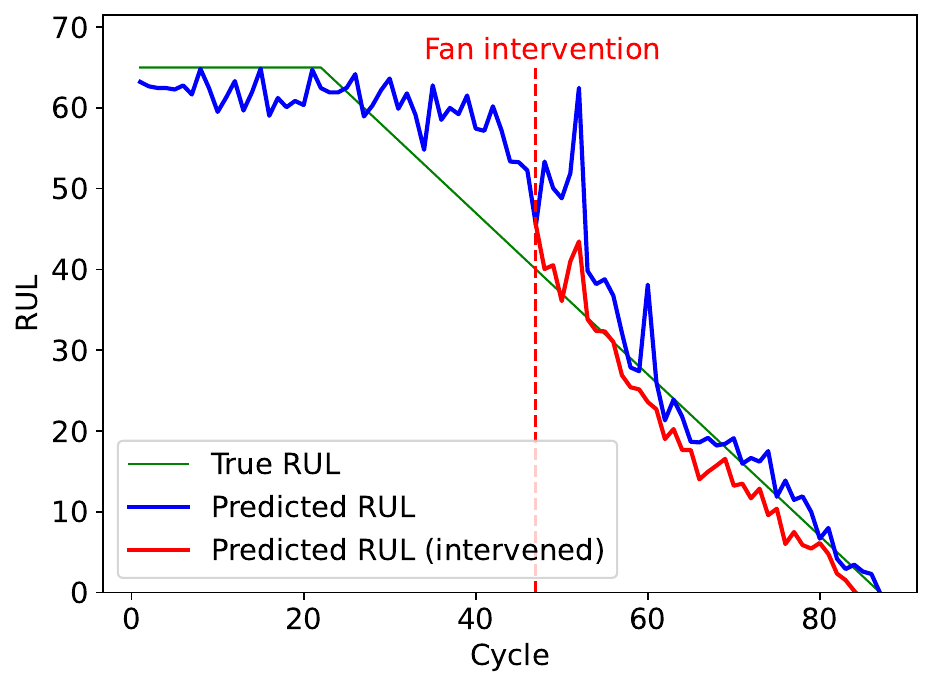}
    \caption{DS04 unit 7}
    \label{fig:interventions:04-7}
    \end{subfigure}
    \begin{subfigure}{0.45\linewidth}
        \includegraphics[width=\linewidth]{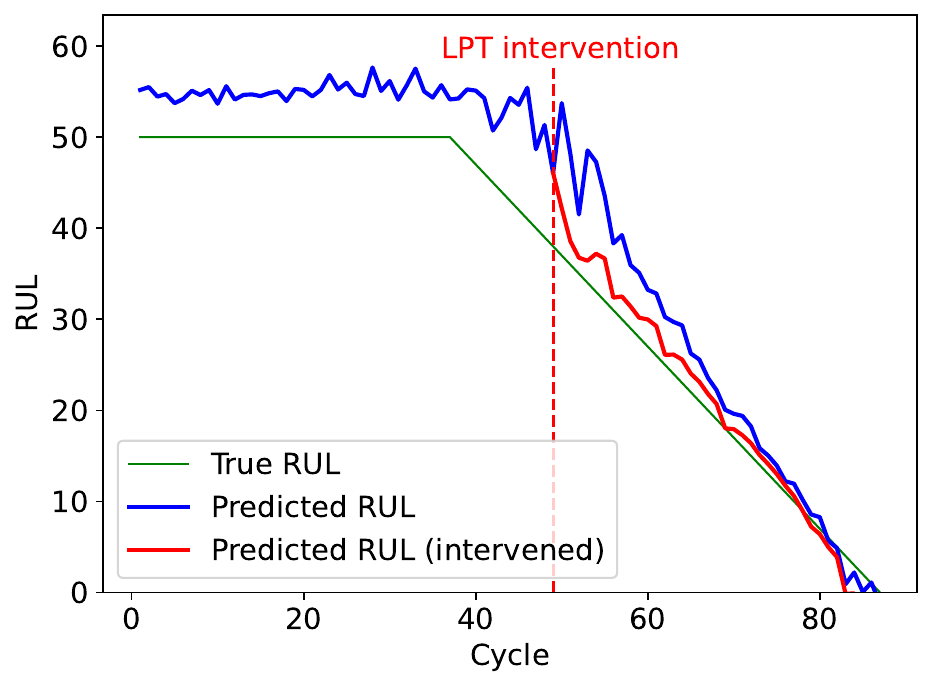}
        \caption{DS07 unit 10}
        \label{fig:interventions:07-10}
    \end{subfigure}
    \caption{Examples of test-time concept interventions.}
    \label{fig:interventions}
\end{figure}
Two examples of interventions are shown Fig.~\ref{fig:interventions}. In both cases, the intervention corrects the overestimated RUL. Generally, interventions will always result in a lower RUL value because the concept activations are increased to 1, which leads to a decreased output RUL.

\begin{table}
    \setlength{\aboverulesep}{0pt}
    \setlength{\belowrulesep}{0pt}
    \centering
    \begin{tabular}{lcc|cc}
        \toprule
         & \multicolumn{2}{c|}{Before interventions} & \multicolumn{2}{c}{After interventions} \\
        Method & RMSE & NASA & RMSE & NASA \\ 
        \midrule
        CNN	& 6.79 & 0.851 & - & - \\
        CNN+CLS	& 6.27 & 0.755 & - & - \\
        \midrule
        Boolean CBM	& 12.50 & 2.334  & 13.59 {\scriptsize\textcolor{red}{+1.09}} & 2.607 {\scriptsize\textcolor{red}{+0.273}} \\
        Fuzzy CBM & 8.08 & 1.168 & 10.87 {\scriptsize\textcolor{red}{+2.79}} & 1.636 {\scriptsize\textcolor{red}{+0.468}} \\
        Hybrid CBM & 6.65 & 0.827 & 6.39 {\scriptsize\textcolor{green}{-0.26}} & 0.761 {\scriptsize\textcolor{green}{-0.066}} \\
        CEM & \textbf{6.15} & \textbf{0.752} & \textbf{5.96} {\scriptsize\textcolor{green}{-0.19}} & \textbf{0.692} {\scriptsize\textcolor{green}{-0.060}} \\
        \bottomrule
    \end{tabular}
    \caption{RUL prediction performance for different methods on Scenario 1 before and after test-time interventions. Average across all test units in DS\{01,04,05,07\}.}
    \label{tab:results-inter-1457}
\end{table}

We also evaluate the impact of interventions quantitatively over all the datasets in Scenario 1, shown in Tab.~\ref{tab:results-inter-1457}. Overall, interventions improve the RUL estimation of the hybrid CBM and the CEM, reducing their RMSE by 0.26 and 0.19 cycles respectively, corresponding to a relative improvement of about 4\%. The improvement is more pronounced in terms of the NASA score, with an 8\% relative improvement, due to the larger penalty given to RUL overestimation in the NASA score's calculation. Interventions do not improve the performance of the Boolean and fuzzy CBM, confirming that these models are not suitable for RUL prediction using a small number of concepts.

\begin{figure}
    \centering
    \includegraphics[width=0.75\linewidth]{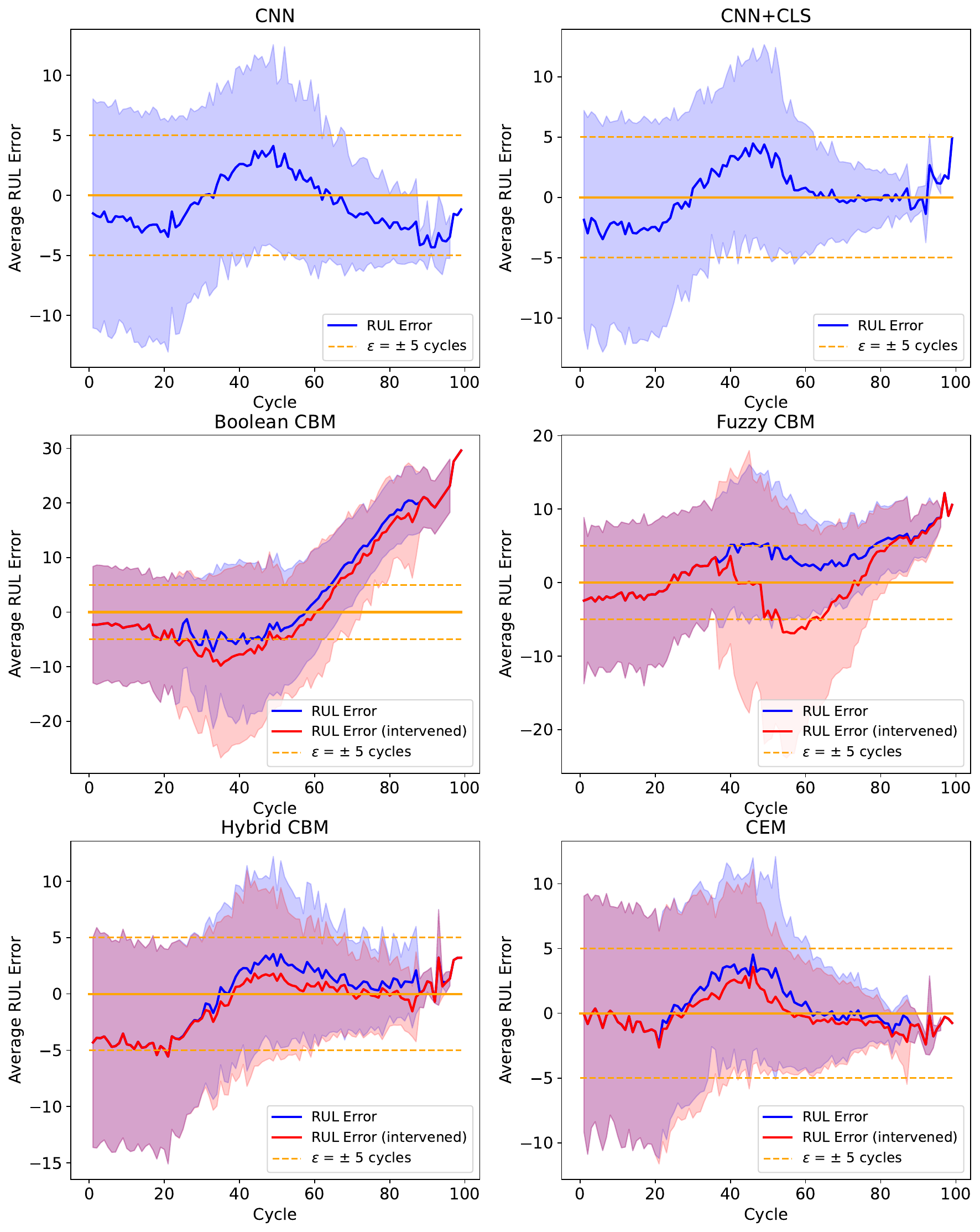}
    \caption{Distribution of RUL prediction errors as a function of cycle number for different methods, before and after test-time interventions. The CEM is more accurate than the CNN and CNN+CLS baselines in the early cycles and towards the end-of-life. Interventions mainly reduce over-estimation of the RUL in the intermediate cycles, when the system transitions from the normal to the abnormal degradation state. Average across all test units in DS\{01,04,05,07\}.}
    \label{fig:rul-error}
\end{figure}

Lastly, we study the distribution of RUL prediction errors as a function of the cycle number in Fig.~\ref{fig:rul-error}. Most predictions lie within an error interval of $\pm 5$ cycles, except the Boolean and fuzzy CBMs, which produce large errors in the later cycles. The CEM is more accurate than the CNN and CNN+CLS baselines, especially in the early cycles and towards the end-of-life. Test-time interventions mainly reduce over-estimation of the RUL in the intermediate cycles when the system transitions from the normal to the abnormal degradation state.

\subsection{Number of concepts used in training}

In real-world cases, labeled data for every degradation mode is usually not available, and the number of potential faults or fault combinations is very large and unknown in advance. Thus, we conduct a study where we vary the number of concepts used during training. The training and testing data is fixed following Scenario 1. We vary the number of concepts from $k=1$ (Fan only) to $k=4$ (Fan, HPC, HPT, LPT), and evaluate all compared methods. For the hybrid CBM, we adapt the extra capacity to match the capacity of the CEM.

\begin{figure}
    \centering
    \begin{subfigure}{0.325\linewidth}
        \includegraphics[width=\linewidth]{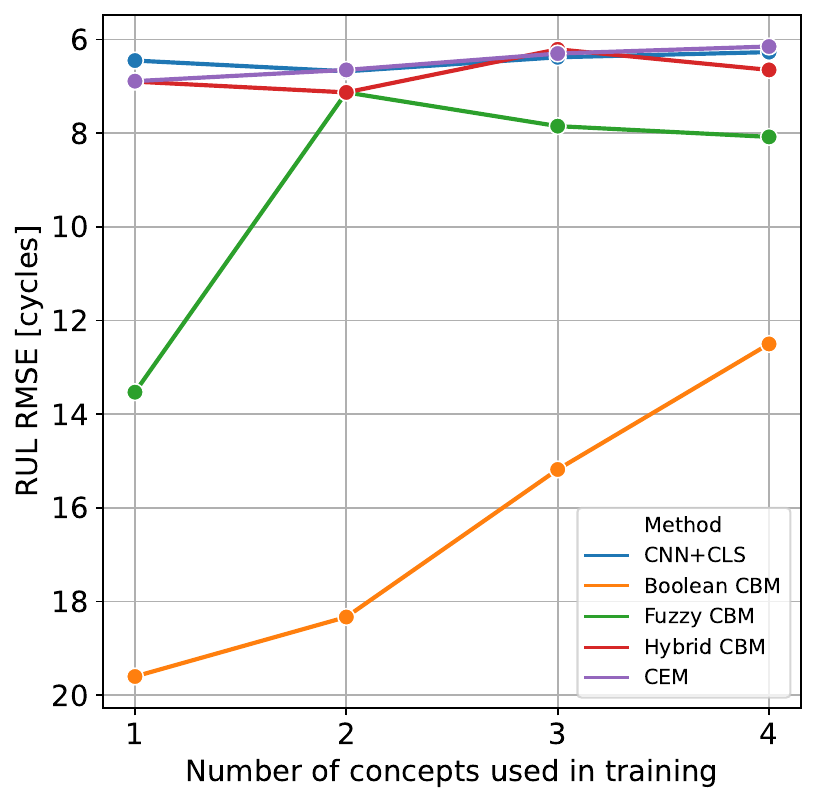}
        \caption{RUL RMSE}
        \label{fig:num-concepts:rmse}
    \end{subfigure}
    \begin{subfigure}{0.325\linewidth}
        \includegraphics[width=\linewidth]{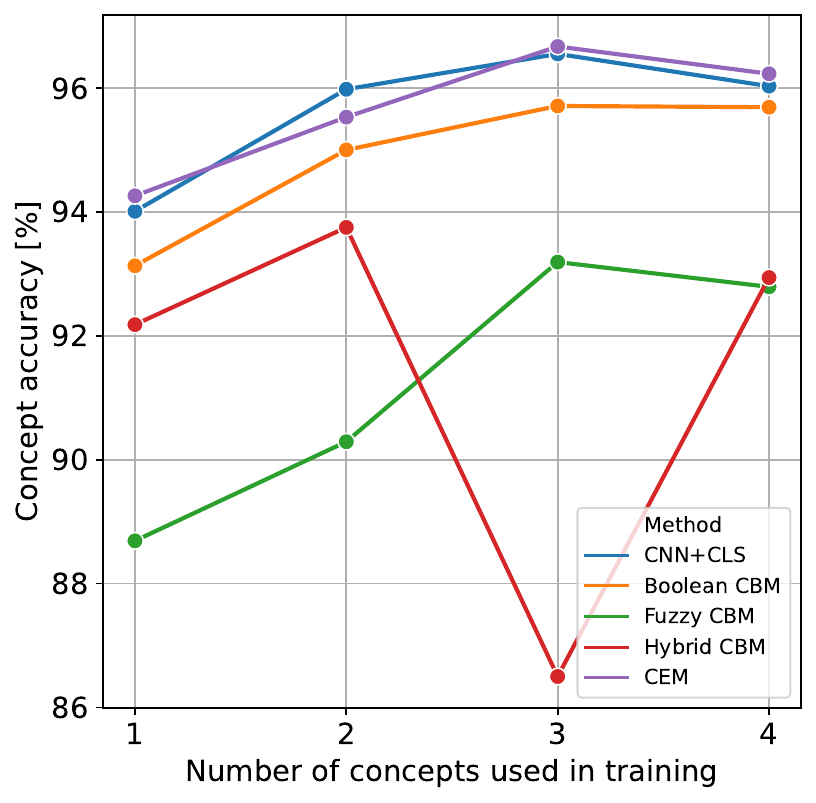}
        \caption{Concept accuracy}
        \label{fig:num-concepts:acc}
    \end{subfigure}
    \begin{subfigure}{0.325\linewidth}
        \includegraphics[width=\linewidth]{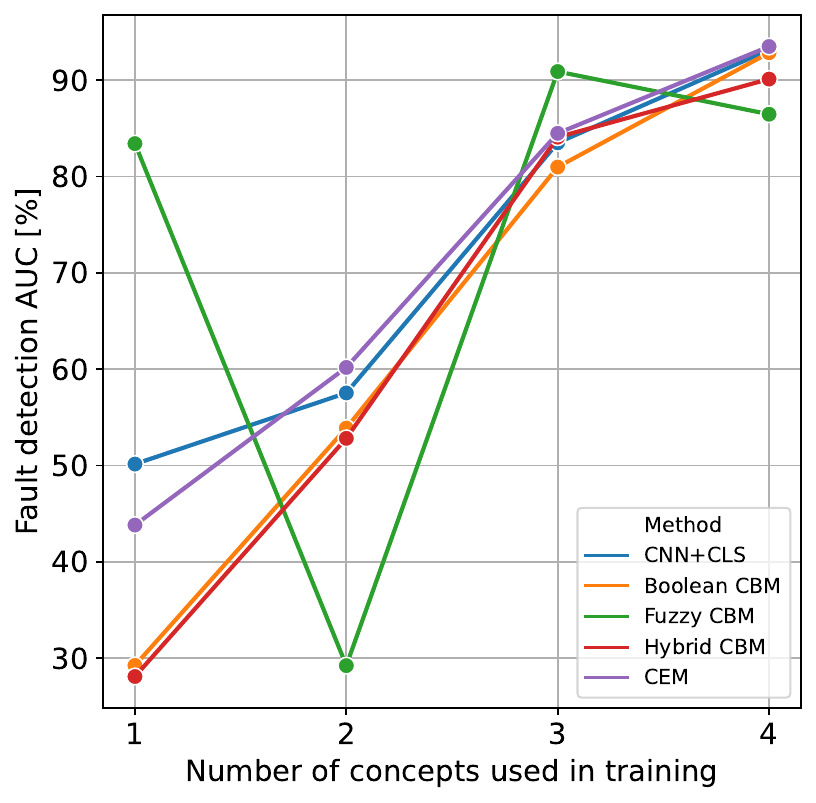}
        \caption{Fault detection AUC}
        \label{fig:num-concepts:auc}
    \end{subfigure}
    \caption{Influence of the number of concepts used for training on RUL RMSE, concept accuracy and fault detection in scenario 1, varying from 1 (Fan only) to 4 (Fan, HPC, HPT, LPT). Average across all test units in DS\{01,04,05,07\}. Performance generally increases when using more concepts. Note that the high fault detection performance of the fuzzy CBM with 1 concept is caused by information leakage between the RUL and the Fan concept.}
    \label{fig:num-concepts}
\end{figure}
The performance in terms of RUL prediction performance (RMSE) and fault detection accuracy (AUC) generally increases when using more concepts (see Fig.~\ref{fig:num-concepts}). The concept accuracy sometimes decreases because the classification task becomes more difficult, especially with HPT and LPT degradations, which have very similar signatures. It is important to note that the hybrid CBM and the CEM are robust to the number of  concepts used and consistently have high RUL prediction performance. This robustness  is explained by the fact that the hybrid CBM can encode RUL information in its additional units, and the CEM can encode RUL information in the negative concept embedding, even when only a single concept is used in the bottleneck.

\begin{figure}
    \centering
    \begin{subfigure}{0.4\linewidth}
        \includegraphics[width=\linewidth]{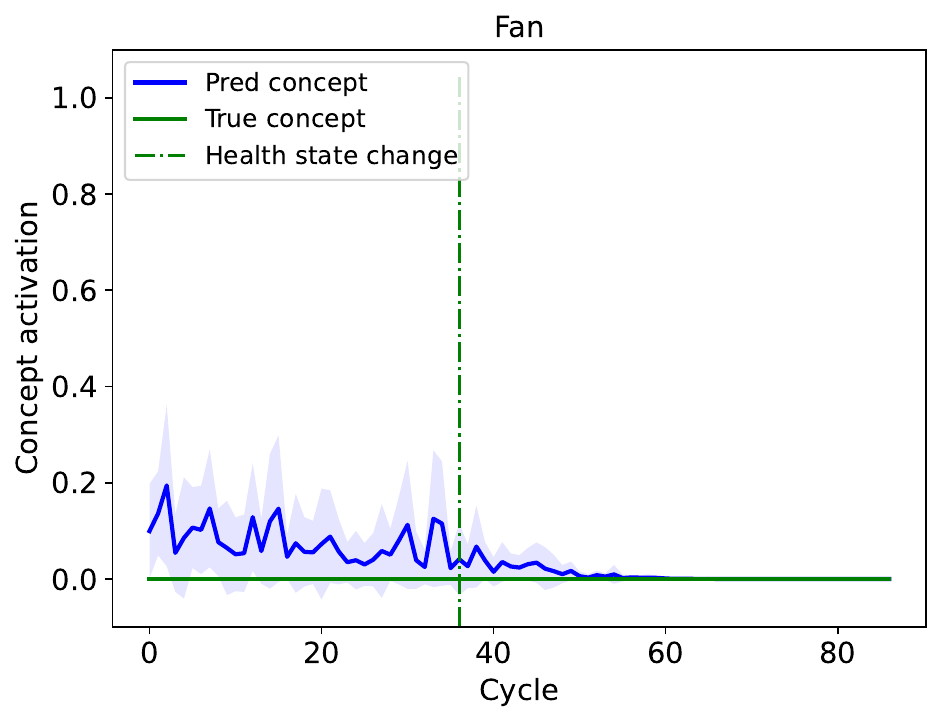}\\
        \includegraphics[width=\linewidth]{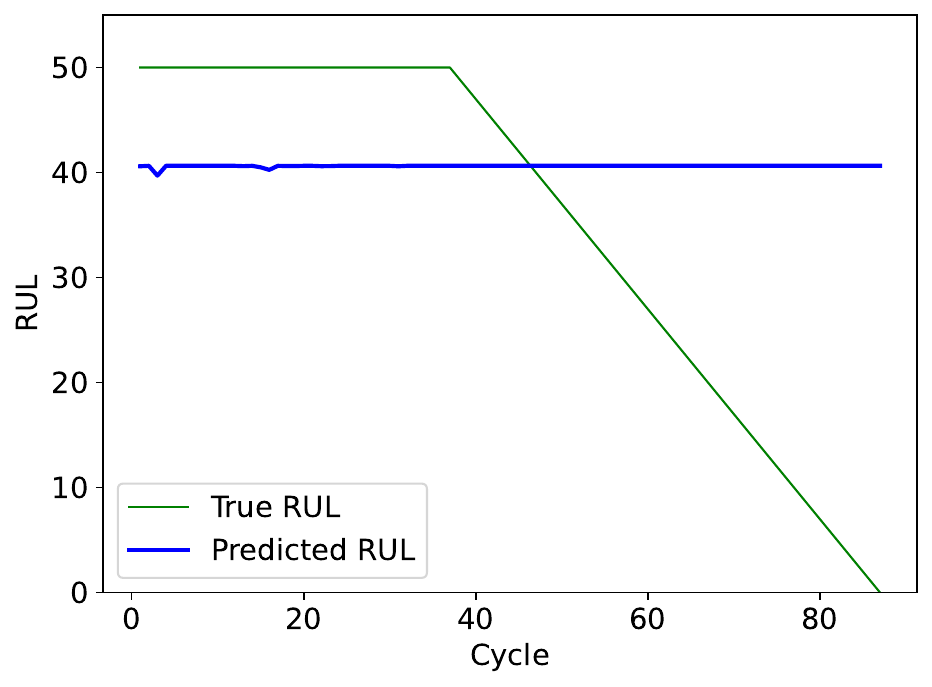}
        \caption{Boolean CBM}
        \label{fig:leakage:cbmbool}
    \end{subfigure}
    \begin{subfigure}{0.4\linewidth}
        \includegraphics[width=\linewidth]{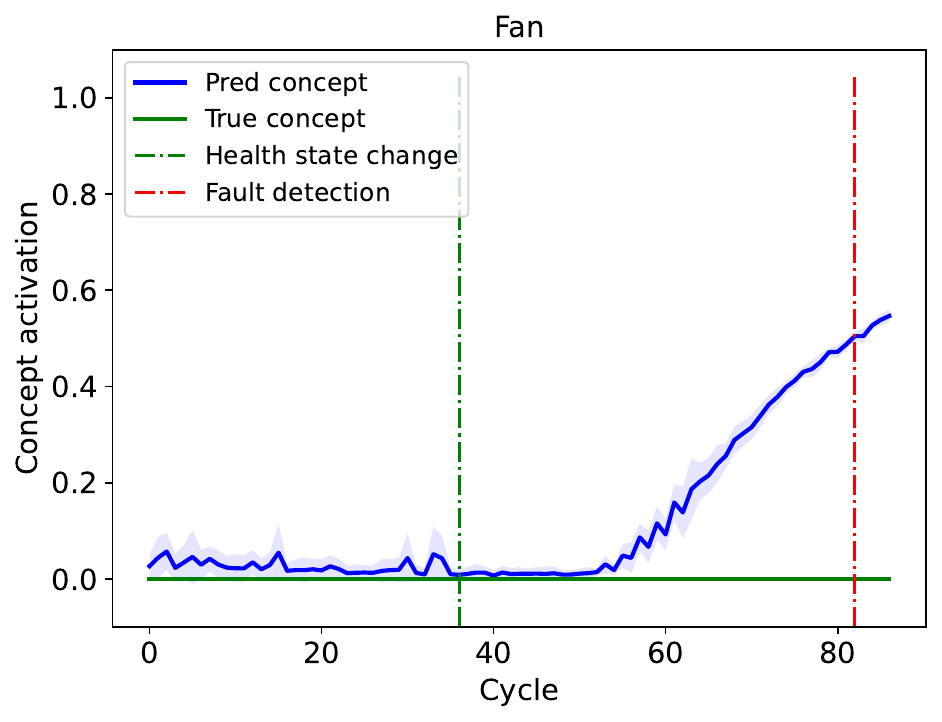}\\
        \includegraphics[width=\linewidth]{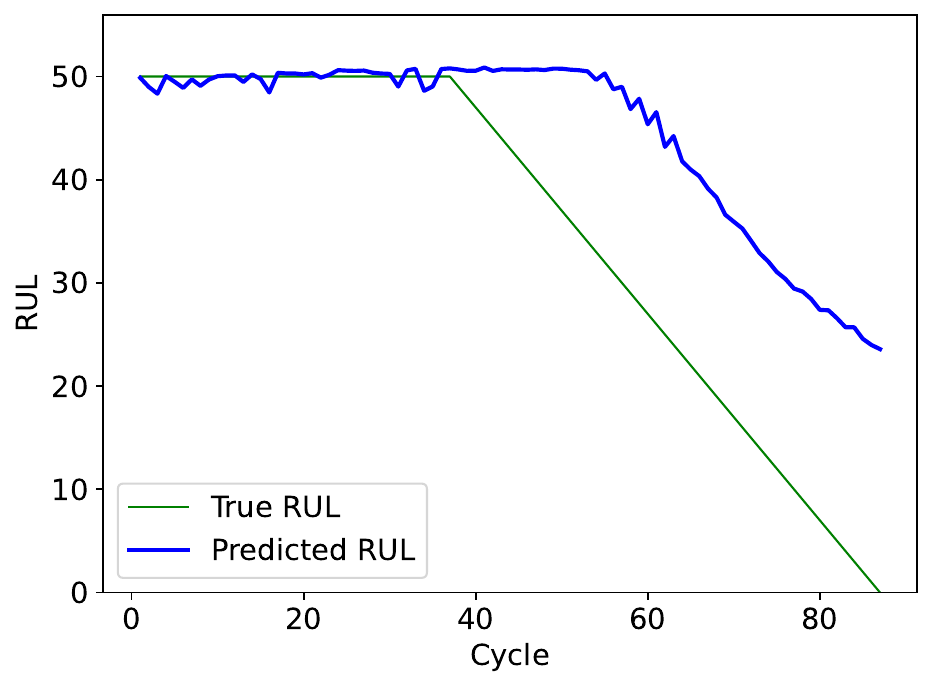}
        \caption{Fuzzy CBM}
        \label{fig:leakage:cbm}
    \end{subfigure}
    \begin{subfigure}{0.4\linewidth}
        \includegraphics[width=\linewidth]{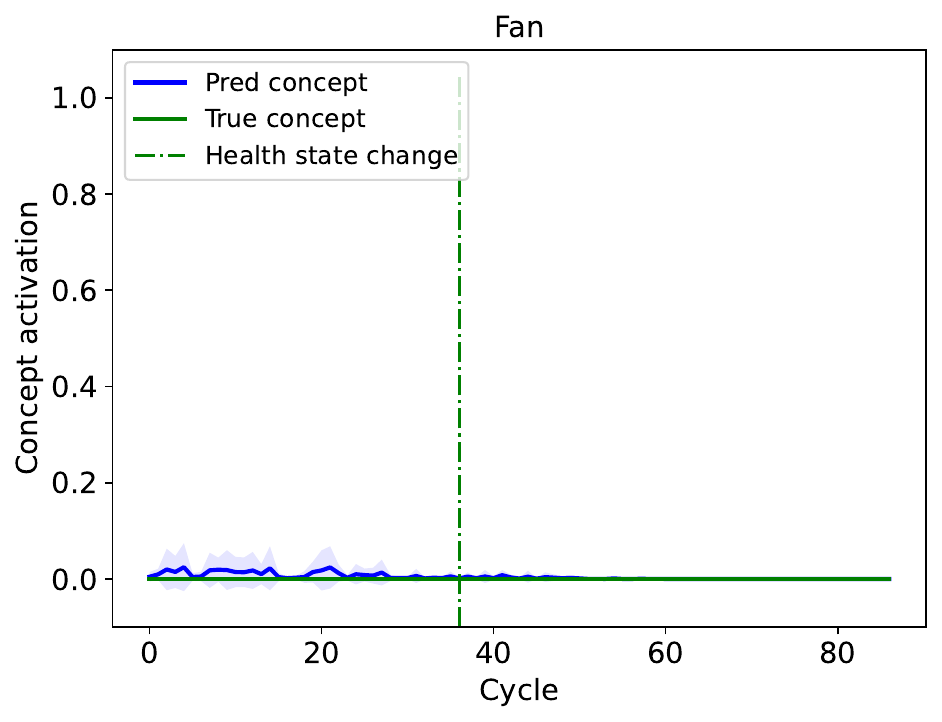}\\
        \includegraphics[width=\linewidth]{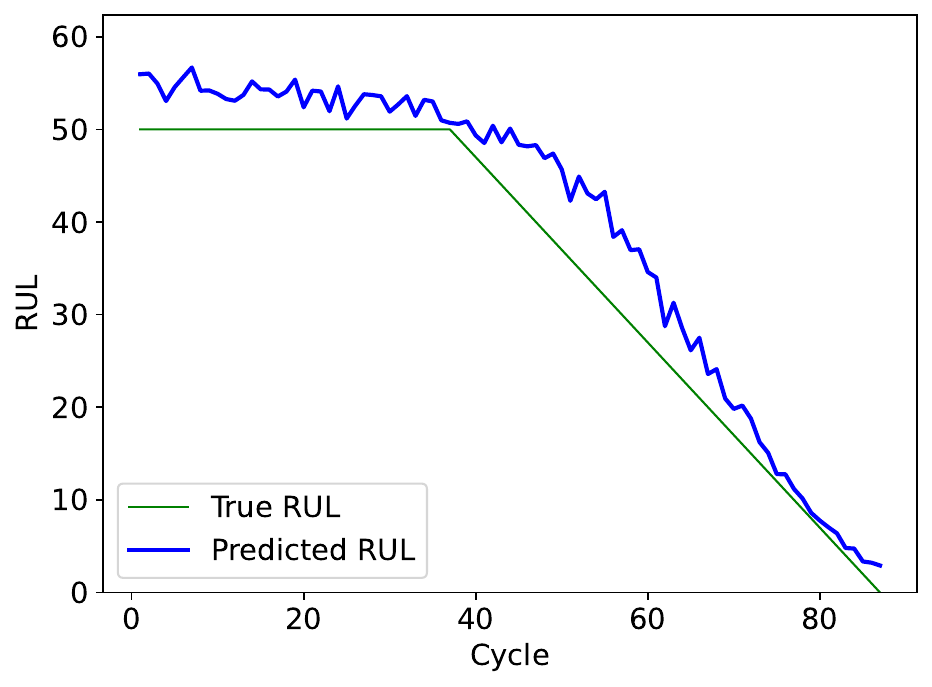}
        \caption{Hybrid CBM}
        \label{fig:leakage:cbmhybrid}
    \end{subfigure}
    \begin{subfigure}{0.4\linewidth}
        \includegraphics[width=\linewidth]{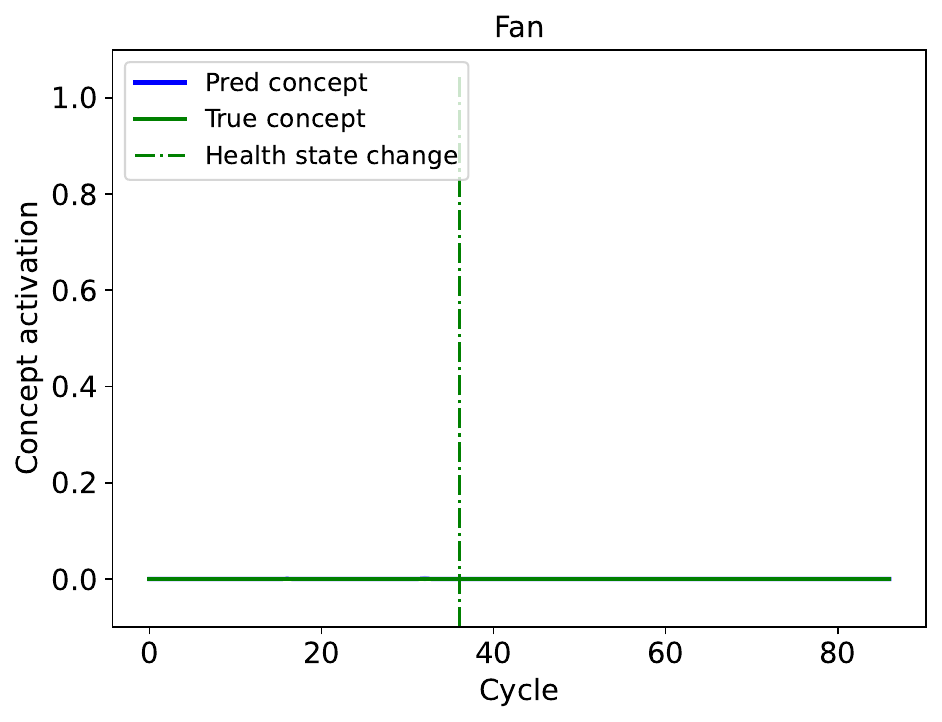}\\
        \includegraphics[width=\linewidth]{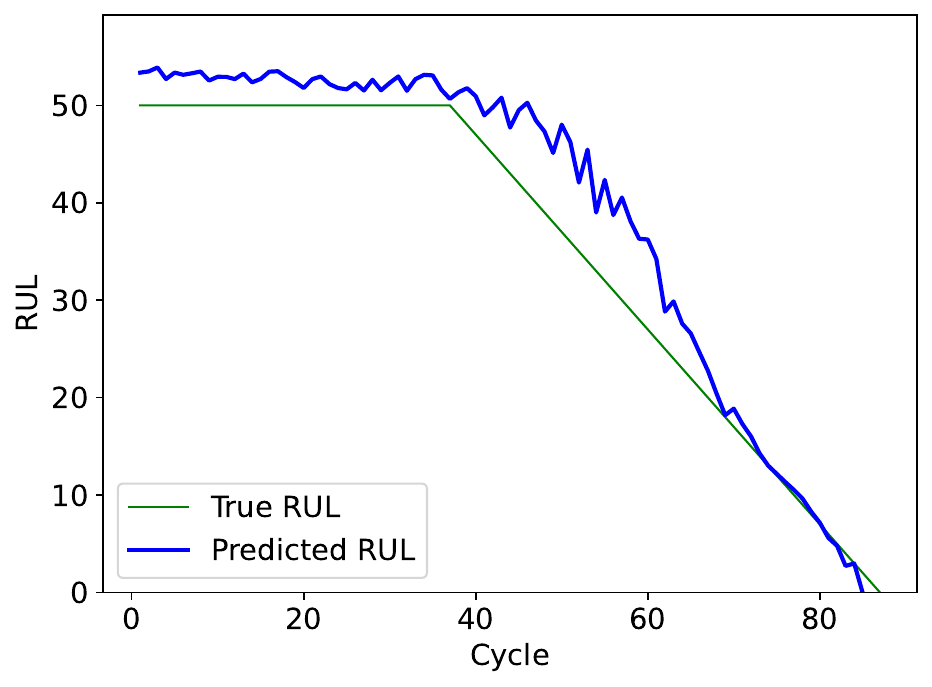}
        \caption{CEM}
        \label{fig:leakage:cem}
    \end{subfigure}
    \caption{Concept activation and predicted RUL when using a single concept (Fan) for unit 10 in DS07 (LPT degradation only). The Boolean and fuzzy CBMs are unable to accurately predict the RUL using a single Boolean or scalar concept. However, information leakage occurs in the fuzzy CBM, with the Fan concept wrongly activating in order to improve the RUL prediction. Hybrid CBM and CEM are accurate without leakage.}
    \label{fig:concepts1000}
\end{figure}
A noteworthy phenomenon is the leakage of information between the task and the concept activations. Information leakage occurs when unintended information contained in concept activations is used by the task predictor. In particular, the soft probabilities in fuzzy CBMs can encode information on the target label \citep{havasi_addressing_2022}. When using only a single fault concept in a CBM, which is insufficient to predict the RUL, the model may encode the linear growth into the concept activation, improving the task loss, while keeping the probability low enough to not hurt the concept loss. This phenomenon occurs, for example, in the CBM trained solely with the Fan concept. In dataset DS07 with LPT degradation only, the Fan concept activation increases linearly from 0 to 0.5, while this concept should not activate at all (see Figure~\ref{fig:leakage:cbm}). As a result, the fuzzy CBM can better predict the output, although it is supposed to output a constant value. Differently, a Boolean CBM cannot leak information on the target as its activations are binary. Thus, the Fan concept does not activate, and the predicted RUL is constant (see Figure~\ref{fig:leakage:cbmbool}). On the contrary, the hybrid CBM and the CEM are able to accurately predict the RUL by encoding it respectively in the extra capacity and in the negative concept embedding, while the Fan concept does not activate, as shown in Figures~\ref{fig:leakage:cbmhybrid} and \ref{fig:leakage:cem}.

\subsection{Latent space visualization}

In the last step of our evaluation, we visualize the latent code space and concept embeddings of the trained CEM model.

\begin{figure}
    \centering
    \begin{subfigure}{0.49\linewidth}
        \includegraphics[width=\linewidth]{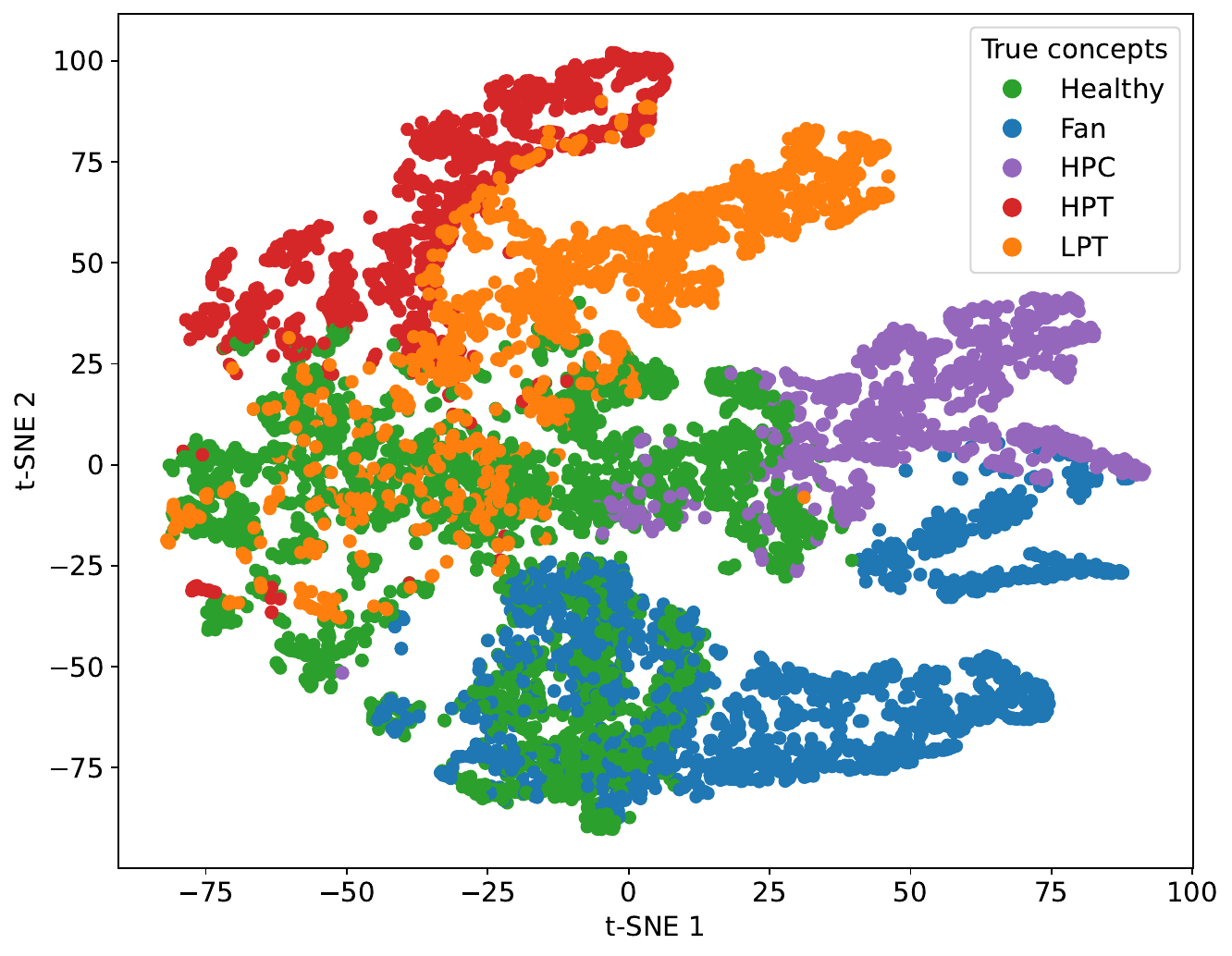}
        \caption{Latent space colored by true concept.}
    \end{subfigure}
    \begin{subfigure}{0.49\linewidth}
        \includegraphics[width=\linewidth]{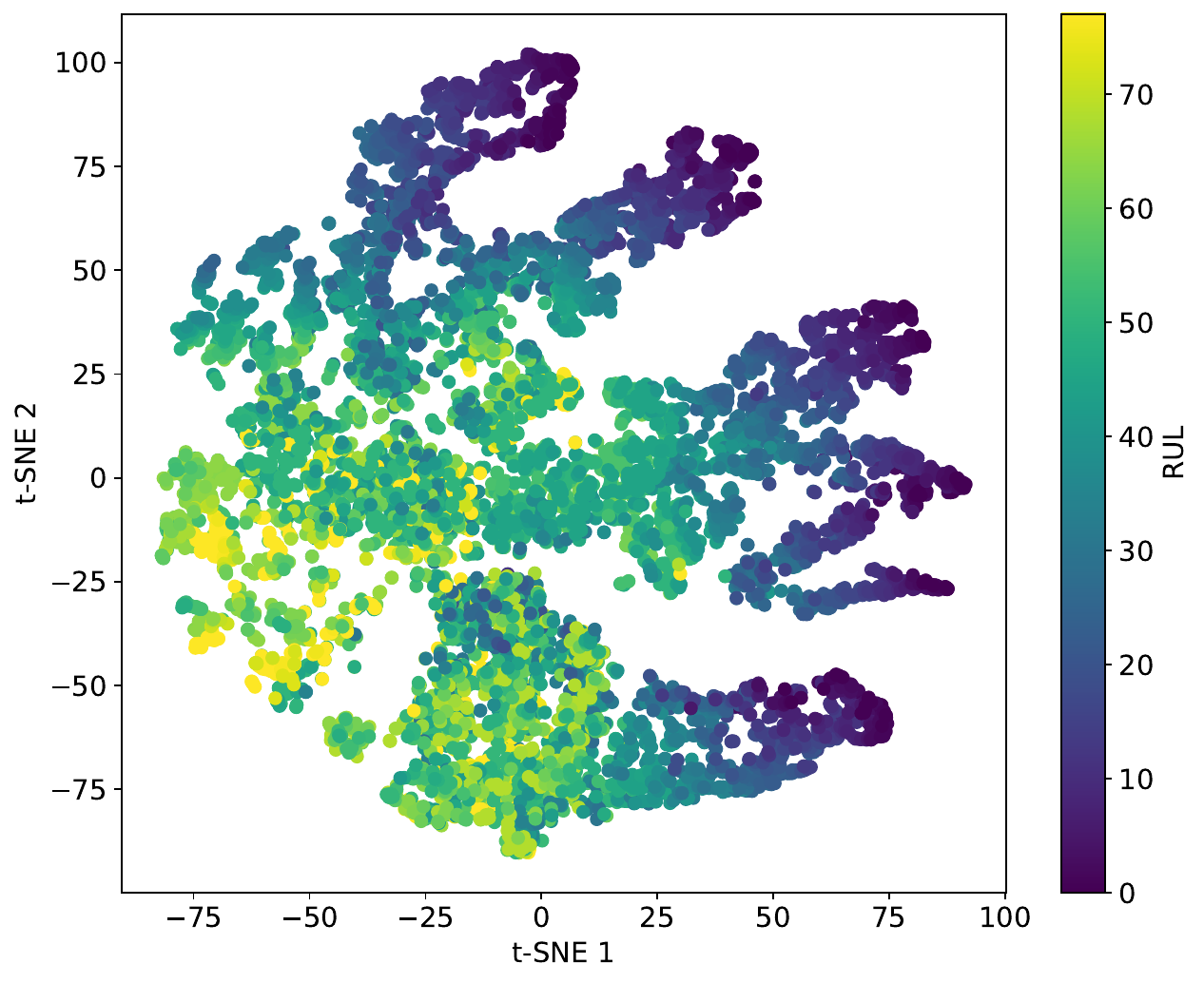}
        \caption{Latent space colored by true RUL value.}
    \end{subfigure}
    \caption{t-SNE visualization of the latent code space of a CEM trained in Scenario 1. Each point represents an input flight window for all test units in datasets DS\{01,04,05,07\}.}
    \label{fig:tsne-latent}
\end{figure}

Fig.~\ref{fig:tsne-latent} visualizes the 256-dimensional latent code ($\mathbf{z}$) using t-SNE \citep{maaten_visualizing_2008}, obtained after the feature extractor. Through joint supervised training with RUL and concept labels, the latent space is structured with respect to both the concepts and the RUL, showing that the CEM captured meaningful degradation semantics. The manifold exhibits four sub-manifolds corresponding to the four degradation modes, with the healthy samples gathering around the center. The RUL value varies smoothly, with high values towards the left and center, and decreasing along the four branches corresponding to the different degradation modes.

\begin{figure}
    \centering
    \begin{subfigure}{\linewidth}
        \includegraphics[width=\linewidth]{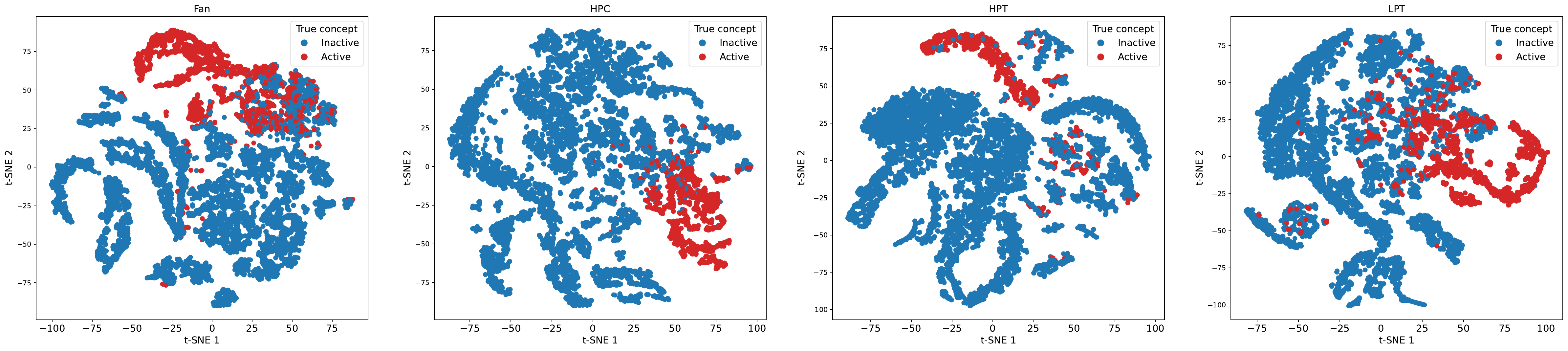}
        \caption{Concept embeddings colored red if active in that sample and blue otherwise.}
    \end{subfigure}
    \begin{subfigure}{\linewidth}
        \includegraphics[width=\linewidth]{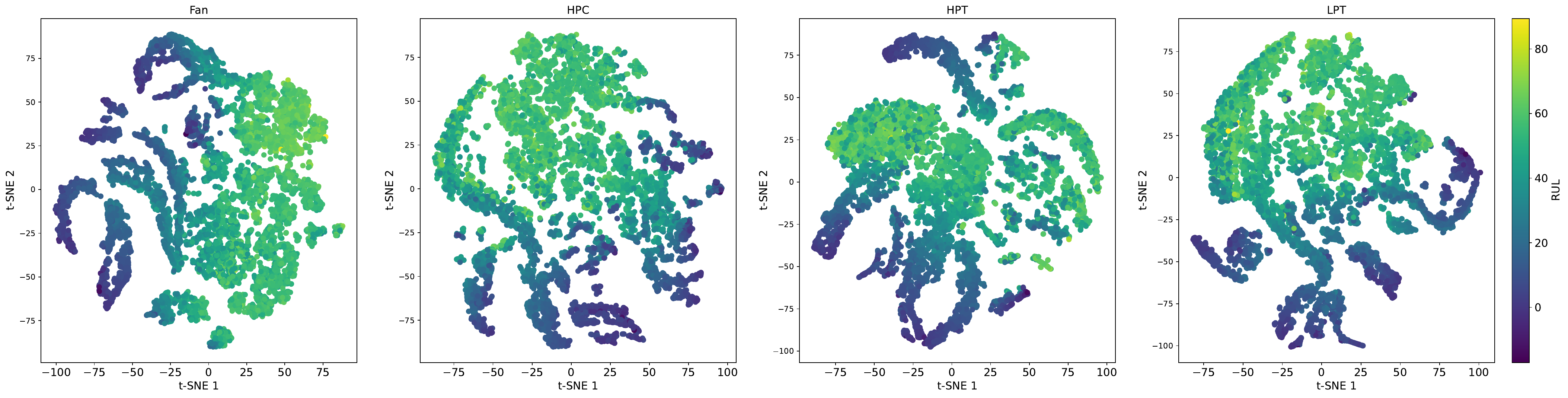}
        \caption{Concept embeddings colored by true RUL value.}
    \end{subfigure}
    \caption{t-SNE visualization of the concept embedding vectors of a CEM trained in Scenario 1. Each point represents an input flight window for all test units in datasets DS\{01,04,05,07\}.}
    \label{fig:tsne-concept-embeddings}
\end{figure}
Lastly, we  visualize the 16-dimensional concept embedding vectors ($\hat{\mathbf{c}}$) corresponding to each concept in Fig.~\ref{fig:tsne-concept-embeddings}. In the upper part, we color the embeddings in red if they are active and blue otherwise for each sample, showing that the learned representations form well-defined clusters. Moreover, the lower part of the figure where each sample is colored by its true RUL value highlights the structure of the embedding space with respect to the RUL, which smoothly varies across the manifold. In particular, both the active and inactive regions are structured, demonstrating that both the negative and positive concept embeddings encode the RUL information.

\section{Discussion}\label{sec:discussion}

\noindent\textbf{Availability of concept labels.} In this work, we defined a concept to be active if the corresponding degradation level exceeds a threshold value. This degradation value is available because we have access to a physics-based performance model that provides this information, allowing us to know the true value. However, the method is not restricted to cases where physics-based performance models are available. In other real-world scenarios, this information could be provided by domain experts who inspect the engine after a failure occurs. Furthermore, only a binary indicator is sufficient, not the precise level of degradation, making it easier to obtain \citep{frusque_semi-supervised_2024}. Note also that concept labels are only used for training, and this information is not available at test time.

\noindent\textbf{Health index methods.} RUL prediction approaches can be either direct or based on a health index (HI) \citep{lei_machinery_2018,yang_remaining_2021,bajarunas_generic_2023}. Direct methods predict the RUL directly from condition monitoring data, which simplifies the process but lacks the ability to monitor  degradation and incorporate domain knowledge. In contrast, HI-based methods use the HI as an intermediate indicator of the system state and then model the mapping from HI to RUL. A downside of HI-based methods is the need for explicit assumptions about the degradation mechanism, which  can be either linear \citep{si_wiener-process-based_2013,yang_health_2016} or non-linear \citep{lu_using_1993,yang_remaining_2021,garmaev_deep_2023}. The present work shares some similarities with HI-based approaches by using intermediate quantities to predict the RUL, but it does not predict an HI or make any assumption about the degradation process.
\section{Conclusion}\label{sec:conclusion}

In this work, we have proposed concept bottleneck models for remaining useful life prediction, using component degradation modes  as human-understandable high-level concepts. These models build  user trust by enforcing the intermediate prediction of concept activations before computing the RUL value and  allowing domain experts to intervene on the concept activations during execution. We  also proposed a test-time intervention strategy adapted for prognostics. Our experiments demonstrated that the concept embedding model (CEM), an extension of CBMs that learns positive and negative embeddings for each concept, was particularly effective among the compared approaches. Both the CEM and the hybrid CBM (which is not fully interpretable) showed superior or comparable performance to standard black-box models without sacrificing interpretability. Thanks to rich concept representations,  these models perform  on par or even better than their non-interpretable counterparts. Moreover, their performance remains robust even when the number of available concepts is very small, which is often  the case in industrial scenarios. 

An important limitation of our proposed approach is the requirement for labeled concept data to predict the RUL. Moreover, there is no guarantee that providing explanations will actually improve human decision-making \citep{alufaisan_does_2020}. Effective  integration into decision support systems used by domain experts will be crucial.

For future work, our methodology can be applied to different neural network architectures, such as recurrent or transformer models, to exploit the sequential nature of the task. Although we  adopted a convolutional network as the base architecture in our work, other architectures can be utilized. Finally, we plan to apply this approach to other industrial systems.

\appendix

\section{Scaling}\label{sec:appendix:scaling}

\begin{table}[H]
    \centering
    \begin{tabular}{lcccc}
        \toprule
        Method & Scaling & RUL RMSE & NASA score & Concept accuracy (\%)\\
        \midrule
        CEM & min-max & 6.70 & 0.879 & 96.07\\
        CEM & standard & \textbf{6.15} & \textbf{0.752} & \textbf{96.23}\\
        \bottomrule
    \end{tabular}
    \caption{Performance of CEM using min-max scaling or standard scaling of input variables (datasets DS01, 04, 05, 07).}
    \label{tab:scaling}
\end{table}

\section{Detailed results}\label{sec:appendix:results}

\begin{table}[H]
    \setlength{\tabcolsep}{2pt}
    \centering
    \resizebox{\linewidth}{!}{
    \begin{tabular}{lccccccccccccccccc}
        \toprule
        \multirow{2}*{Method} & \multicolumn{4}{c}{DS01} & \multicolumn{4}{c}{DS04} & \multicolumn{4}{c}{DS05} & \multicolumn{4}{c}{DS07} & \multirow{2}*{Average} \\
         & 7 & 8 & 9 & 10 & 7 & 8 & 9 & 10 & 7 & 8 & 9 & 10 & 7 & 8 & 9 & 10 \\
        \midrule
        CNN         &  \textbf{2.72} &   7.56 &   7.09 &   7.49 &   9.24 &  10.87 &  10.82 &   5.82 &   6.48 &   2.49 &  \textbf{2.94} &   4.40 &   3.05 &   9.06 &  12.49 &  6.12 &   6.79 \\
        CNN+CLS     &  3.26 &   4.68 &   8.64 &   4.21 &   \textbf{6.40} &   9.99 &   \textbf{9.33} &   5.05 &   6.87 &   \textbf{1.70} &  3.32 &   5.30 &   2.87 &   8.69 &  12.93 &  7.02 &   6.27 \\
        \midrule
        Boolean CBM &  9.36 &  14.44 &  12.05 &  14.06 &  15.02 &  15.19 &  15.94 &  13.26 &  11.58 &  10.91 &  9.43 &  11.37 &  10.20 &   9.46 &  18.80 &  8.90 &  12.50 \\
        Fuzzy CBM   &  5.16 &   5.71 &  10.69 &   5.09 &  12.90 &  12.20 &  10.27 &   7.66 &   \textbf{4.65} &   2.69 &  6.72 &   7.62 &   4.85 &  11.95 &  \textbf{11.89} &  9.15 &   8.08 \\
        Hybrid CBM  &  2.02 &   7.35 &   \textbf{6.12} &   5.55 &   9.04 &  13.07 &  10.76 &   7.00 &   7.52 &   4.14 &  3.57 &   \textbf{2.35} &   2.91 &   \textbf{7.22} &  12.84 &  \textbf{4.98} &   6.65 \\
        CEM    &  4.29 &  \textbf{3.65} &  10.08 &  \textbf{3.56} &  6.68 &  \textbf{7.91} &  10.77 &  \textbf{4.23} &  6.59 &  2.75 &  3.80 &  4.50 &  \textbf{2.75} &  8.14 &  12.26 &  6.45 &  \textbf{6.15} \\
         \bottomrule
    \end{tabular}
    }
    \caption{RUL RMSE in cycles for different methods on Scenario 1.}
    \label{tab:results-rmse-1457}
\end{table}

\begin{table}[H]
    \setlength{\tabcolsep}{2pt}
    \centering
    \resizebox{\linewidth}{!}{
    \begin{tabular}{lccccccccccccc}
        \toprule
        \multirow{2}*{Method} & \multicolumn{4}{c}{DS01} & \multicolumn{4}{c}{DS03} & \multicolumn{4}{c}{DS07} & \multirow{2}*{Average} \\
         & 7 & 8 & 9 & 10 & 7 & 8 & 9 & 10 & 7 & 8 & 9 & 10 \\
        \midrule
        CNN         &  \textbf{2.75} &   7.60 &  \textbf{5.91} &   7.12 &   8.77 &   5.17 &  5.49 &   \textbf{5.55} &   \textbf{2.95} &   7.49 &  \textbf{13.27} &  \textbf{4.85} &   6.41 \\
        CNN+CLS     &  3.87 &   \textbf{4.46} &  8.31 &   \textbf{3.71} &   \textbf{6.78} &   \textbf{4.83} &  4.49 &   7.62 &   3.60 &   7.64 &  13.39 &  6.11 &   \textbf{6.23} \\
        \midrule
        Boolean CBM &  8.42 &  14.06 &  8.41 &  13.96 &  14.84 &  10.66 &  7.98 &  10.28 &  10.14 &  10.76 &  18.25 &  6.59 &  11.20 \\
        Fuzzy CBM   &  3.91 &   8.27 &  7.07 &   7.77 &  10.99 &   9.03 &  8.43 &   7.55 &   5.53 &   8.46 &  15.51 &  5.01 &   8.13 \\
        Hybrid CBM  &  3.50 &   5.51 &  7.46 &   5.22 &   7.20 &   6.34 &  4.81 &   6.19 &   3.54 &   \textbf{7.24} &  14.91 &  5.55 &   6.46 \\
        CEM         &  3.64 &   4.90 &  7.93 &   4.84 &   \textbf{6.78} &   5.47 &  \textbf{3.83} &   6.94 &   4.02 &   7.61 &  14.20 &  6.33 &   6.37 \\
         \bottomrule
    \end{tabular}
    }
    \caption{RUL RMSE in cycles for different methods on Scenario 2.}
    \label{tab:results-rmse-137}
\end{table}

 \bibliographystyle{elsarticle-harv} 
 \bibliography{references}





\end{document}